\documentclass[10pt,journal,compsoc]{IEEEtran}
\ifCLASSINFOpdf
\else
\fi

\usepackage{blindtext}
\usepackage{graphicx}
\usepackage{algorithmic}
\usepackage{array}
\usepackage{subfigure}
\usepackage{epstopdf}
\usepackage{url}
\usepackage{hyperref}
\usepackage{algorithm}
\usepackage{algorithmic}
\usepackage{lipsum}
\usepackage{amsmath,amsfonts,amsthm,amssymb}
\usepackage[numbers]{natbib}
\usepackage{color}

\newcommand{\bbeta}{\boldsymbol{\eta}}
\newcommand{\bbsigma}{\boldsymbol{\sigma}}
\newcommand{\bbtheta}{\boldsymbol{\theta}}
\newcommand{\bbphi}{\boldsymbol{\phi}}
\newcommand{\bbbeta}{\boldsymbol{\beta}}
\newcommand{\bbalpha}{\boldsymbol{\alpha}}
\newcommand{\bbxi}{\boldsymbol{\xi}}

\newcommand{\bblambda}{\boldsymbol{\lambda}}
\newcommand{\bbomega}{\boldsymbol{\omega}}
\newcommand{\bbmu}{\boldsymbol{\mu}}
\newcommand{\bbepsilon}{\boldsymbol{\epsilon}}
\newcommand{\bbzero}{\boldsymbol{0}}
\newcommand{\bbI}{\mathbf{I}}
\newcommand{\bbf}{\mathbf{f}}
\newcommand{\bbg}{\mathbf{g}}
\newcommand{\bbx}{\mathbf{x}}
\newcommand{\bbX}{\mathbf{X}}
\newcommand{\bbY}{\mathbf{Y}}

\newcommand{\ytilde}{\tilde{y}}
\newcommand{\bbz}{\mathbf{z}}
\newcommand{\bbZ}{\mathbf{Z}}
\newcommand{\ep}{\mathbb{E}}
\newcommand{\KL}{\mathbf{KL}}

\newcommand{\mD}{\mathcal{D}}
\newcommand{\mU}{\mathcal{U}}

\newcommand{\argmax}{\operatornamewithlimits{argmax}}

\begin{document}
%
\title{Max-Margin Deep Generative Models for (Semi-)Supervised Learning}
%
%
%
%

\author{Chongxuan~Li,
        Jun~Zhu,~\IEEEmembership{Member, IEEE,}
        and~Bo~Zhang
\IEEEcompsocitemizethanks{
\IEEEcompsocthanksitem C. Li, J. Zhu and B. Zhang are with Department of Computer Science
and Technology; TNList Lab; State Key Laboratory for Intelligent
Technology and Systems; Center for Bio-Inspired Computing
Research, Tsinghua University, Beijing, 100084 China. Email: chongxuanli1991@gmail.com; dcszj@tsinghua.edu.cn;
dcszb@tsinghua.edu.cn.}
\thanks{Manuscript received April 19, 2005; revised August 26, 2015.}}

%
%

\markboth{Journal of \LaTeX\ Class Files,~Vol.~14, No.~8, August~2015}%
{Shell \MakeLowercase{\textit{et al.}}: Bare Demo of IEEEtran.cls for Computer Society Journals}
%



\IEEEtitleabstractindextext{%
\begin{abstract}
Deep generative models (DGMs) are effective on learning multilayered representations
of complex data and performing inference of input data by exploring the generative
ability. However,
it is relatively insufficient to 
empower the discriminative ability of DGMs on making accurate predictions. 
This paper presents max-margin deep
generative models (mmDGMs) and a class-conditional variant (mmDCGMs), which explore the strongly discriminative principle of max-margin
learning to improve the predictive performance of DGMs in both supervised and semi-supervised learning, while retaining the generative
capability. In semi-supervised learning, we use the predictions of a max-margin classifier as the missing labels instead of performing full posterior inference for efficiency; we also introduce additional max-margin and label-balance regularization terms of unlabeled data for effectiveness. We develop an efficient doubly stochastic subgradient algorithm for the piecewise linear objectives in different settings.
Empirical results on various datasets demonstrate that: (1) max-margin learning
can significantly improve the prediction performance of DGMs and meanwhile
retain the generative ability; (2) in supervised learning, mmDGMs are competitive to the best
fully discriminative networks when employing convolutional neural networks as the generative and recognition models; and (3) in semi-supervised learning, mmDCGMs can perform efficient inference and achieve state-of-the-art classification results on several benchmarks.
\end{abstract}

}

\maketitle

\IEEEdisplaynontitleabstractindextext

%
\IEEEpeerreviewmaketitle

\IEEEraisesectionheading{\section{Introduction}\label{sec:introduction}}

%
%
%
%
\IEEEPARstart{M}{ax-margin} learning has been effective on learning discriminative models, with many examples such as
univariate-output support vector machines (SVMs)~\cite{Cortes:1995} and multivariate-output max-margin Markov networks (or structured SVMs)~\cite{Taskar:03,Altun:03,Tsochantaridis:04}.
However, the ever-increasing size of complex data makes it hard to construct such a fully discriminative model, which has only a single layer
of adjustable weights, due to the facts that:
(1) the manually constructed features may not well capture the underlying high-order statistics; and (2) a fully discriminative approach cannot
reconstruct the input data when noise or missing values are present.

To address the first challenge, previous work has considered incorporating latent variables into a max-margin model, including
partially observed maximum entropy discrimination Markov networks~\cite{Zhu:08b}, structured latent SVMs~\cite{Yu:2009} and max-margin min-entropy models~\cite{Miller:2012}.
All this work has primarily focused on a shallow structure of latent variables. To
improve the flexibility, learning SVMs with a deep latent structure has been presented in~\cite{Tang:2013}. However, these methods do not
address the second challenge, which requires a generative model to describe the inputs. The recent work on learning max-margin generative models includes
max-margin topic models~\cite{zhu12jmlr,zhu14jmlr-lda}, max-margin Harmoniums~\cite{Chen:2012pami},
and nonparametric Bayesian latent SVMs~\cite{zhu14jmlr} which can infer the dimension of latent features from data.
However, these methods only consider the shallow structure of latent variables, which may not be flexible enough to describe complex data.

Much work has been done on learning generative models with a deep structure of nonlinear hidden variables, including deep belief networks~\cite{Salakhutdinov:09,Lee:09,Ranzato:11}, autoregressive models~\cite{Larochelle:11,Gregor:14}, stochastic variations of autoencoders ~\cite{vincent2010stacked,bengio2013generalized,Bengio:14}
and Generative Adversarial Nets (GANs)~\cite{goodfellow:14,radford2015unsupervised}. For such models, inference is a challenging problem, which has motivated much recent progress on stochastic variational inference algorithms~\cite{kingma14iclr,danilo14icml,bornschein2014reweighted,burda2015importance}.
However, the primary focus of deep generative models (DGMs) has been on unsupervised learning, with the goals of learning latent
representations and generating input samples. Though the latent representations can be used with a downstream classifier to make predictions,
it is often beneficial to learn a joint model that considers both input and response variables.
The recent work on semi-supervised deep generative models~\cite{kingma14nips,springenberg16,maaloe16,salimans2016improved} proves the effectiveness of DGMs on modeling the density of unlabeled data to benefit the prediction task (See Sec.~\ref{sec:related_work} for a detailed discussion). However, it remains open whether the discriminative max-margin learning is suitable for this task.

In this paper, we revisit the max-margin principle and present max-margin deep generative models (mmDGMs), which learn
multilayered representations that are good for both classification and input inference. Our mmDGMs conjoin the flexibility of DGMs on describing input data and the strong discriminative ability of max-margin learning on making accurate predictions.
Given fully labeled data, we formulate mmDGMs as solving a variational inference problem of a DGM regularized by a set of max-margin posterior constraints,
which bias the model to learn representations that are good for prediction. We define the max-margin posterior
constraints as a linear functional of the target variational distribution of the latent presentations. To optimize the joint learning problems, we develop a doubly stochastic subgradient descent algorithm, which generalizes the
Pagesos algorithm~\cite{shai11pegasos} to consider nontrivial latent variables.
For the variational distribution, we build a recognition model to capture the nonlinearity, similar
as in~\cite{kingma14iclr,danilo14icml}.

To reduce the dependency on fully labeled data, we further propose a class-conditional variant of mmDGMs (mmDCGMs) to deal with partially labeled data for semi-supervised learning, where the amount of unlabeled data is typically much larger than that of labeled ones. Specifically, mmDCGMs employ a deep max-margin classifier to infer the missing labels for unlabeled data and a class-conditional deep generative model~\cite{kingma14nips} to capture the joint distribution of the data, labels and latent variables. Unlike~\cite{rasmus15,springenberg16,salimans2016improved}, our mmDCGMs separate the pathways of inferring labels and latent variables completely and can generate images given a specific class.
Instead of inferring the full posterior of labels as in~\cite{kingma14nips,maaloe16}, which is computationally expensive for large datasets, we use the prediction of the classifier as a point estimation of the label to speed-up the training procedure. We further design additional max-margin and label-balance regularization terms of unlabeled data to enhance the classifier and significantly boost the classification performance.

We consider two types of networks used in our mmDGMs and mmDCGMs---multiple layer perceptrons (MLPs) as in~\cite{kingma14iclr,danilo14icml} and
convolutional neural networks (CNNs)~\cite{Lecun:98}.
In the CNN case, following~\cite{Dosovitskiy:2014}, we apply unpooling, convolution and rectification sequentially to form a highly non-trivial deep generative network to generate images from the latent variables that are learned automatically by a recognition model using a standard CNN. We present the detailed network structures in the experiment section.
Empirical results on the widely used MNIST~\cite{Lecun:98}, SVHN~\cite{Netzer:11} and small NORB~\cite{lecun2004learning} datasets demonstrate that: (1) mmDGMs can significantly improve the prediction performance in supervised learning, which is competitive to the best feedforward neural networks,
while retaining the capability of generating input samples and completing their missing values; and (2) mmDCGMs can achieve state-of-the-art classification results with efficient inference and disentangle styles and classes based on raw images in semi-supervised learning.

In summary, our main contributions are:
\begin{itemize}
\item We present max-margin DGMs for both supervised and semi-supervised settings to significantly enhance the
discriminative power of DGMs while retaining their generative ability;
\item We develop efficient algorithms to solve the joint learning problems, which involve intractable expectation and non-smooth piecewise linear operations;
\item We achieve state-of-the-art results on several benchmarks in semi-supervised learning and competitive prediction accuracy as the fully discriminative CNNs in supervised learning. 
\end{itemize}
The rest of the paper is structured as follows. Section 2 surveys the related work. Section 3 presents max-margin deep generative models for both supervised and semi-supervised learning. Section 4 presents experimental results. Finally, Section 5 concludes.

\section{Related Work}
\label{sec:related_work}

Deep generative models (DGMs) are good at discovering the underlying structures in the input data, but the training of the model parameters and inference of the posterior distribution are highly nontrivial tasks. Recently, significant progress has been made on enriching the representative power of variational inference and Markov chain Monte Carlo methods for posterior inference, such as variational Autoencoders (VAEs)~\cite{kingma14iclr,danilo14icml} and neural adaptive MCMC~\cite{du2015learning}.
VAEs~\cite{kingma14iclr,danilo14icml} build a recognition model to infer the posterior of latent variables and the parameters are trained to optimize a variational bound of the data likelihood. Neural adaptive MCMC~\cite{du2015learning} employs a similar recognition model as the proposal distribution for importance sampling to estimate the gradient of log-posterior and hence can perform approximate Bayesian inference of DGMs.

To learn the parameters, besides the commonly used MLE estimator as adopted by VAEs, recent work has proposed various objectives. For example, Generative Adversarial Nets (GANs)~\cite{goodfellow:14} construct a discriminator to distinguish the generated samples from the training data and the parameters are trained based on a minimax two-player game framework. Generative Moment Matching Networks (GMMNs)~\cite{li2015generative,dziugaite2015training} generate samples from a directed deep generative model, which is trained to match all orders of statistics between training data and samples from the model. The very recent work~\cite{yong2016conditional} extends the ideas to learn conditional GMMNs with much broader applicability. 

Extensive work has been focusing on realistic image generation in unsupervised setting. For example, DRAW~\cite{gregor2015draw} employs recurrent neural networks as the generative model and recognition model and introduces a 2-D attention mechanism to generate sequences of real digits step by step. MEM-VAE~\cite{li2016learning} leverages an external memory and an attention mechanism to encode and retrieve the detailed information lost in the recognition model to enhance DGMs.
LAP-GAN~\cite{denton2015deep} proposes a
cascade of GANs to generate high quality natural images through a Laplacian pyramid framework~\cite{burt1983laplacian}.
DCGAN~\cite{radford2015unsupervised} adopts fractionally strided convolution networks in the generator to learn the spatial upsampling and refines the generated samples.

Some recent advances~\cite{kingma14nips,maaloe16,springenberg16,salimans2016improved,rasmus15} have been made on extending DGMs 
to deal with partially observed data.
For example, the conditional VAEs~\cite{kingma14nips} treat labels as conditions of DGMs to describe input data; they perform posterior inference of labels given unlabeled data and can generate a specific class of images. ADGM~\cite{maaloe16} introduces auxiliary latent variables to DGMs to make the variational distribution more expressive and does well in semi-supervised learning. Cat-GAN~\cite{springenberg16} generalizes GANs with a categorical discriminative network and an objective function that includes the mutual information between the input data and the prediction of the discriminative network.~\cite{salimans2016improved} proposes feature mapping, virtual batch normalization and other techniques to improve the performance of GANs on semi-supervised learning and image generation.
The Ladder Network~\cite{rasmus15} achieves excellent classification results in semi-supervised learning by employing lateral connections between autoencoders to reduce the competition between the invariant feature extraction and the reconstruction of object details.

Our work is complimentary to the above progress in the sense that we investigate a new criterion (i.e., max-margin learning) for DGMs in both supervised and semi-supervised settings. Some preliminary results on the fully supervised mmDGMs were published in~\cite{li2015max}, while the semi-supervised extensions are novel.

\section{Max-margin Deep Generative Models}

We now present the max-margin deep generative models for supervised learning and their class-conditional variants for  semi-supervised learning. For both methods, we present efficient algorithms.

\subsection{Basics of Deep Generative Models}

We start from a general setting,
where we have $N$ i.i.d. data $\bbX = \{ \bbx_n \}^{N}_{n = 1}$.
A deep generative model (DGM) assumes that
each $\bbx_n \in \mathbb{R}^D$ is generated from a vector of latent variables $\bbz_n \in \mathbb{R}^K$,
which itself follows some distribution. The joint probability of a DGM is as follows:
\begin{equation}\label{eq:DGM-joint-dist}
p(\bbX, \bbZ| \bbalpha, \bbbeta)  = \prod^{N}_{n = 1} p(\bbz_n | \bbalpha) p(\bbx_n | \bbz_n, \bbbeta),
\end{equation}
where $p(\bbz_n | \bbalpha)$ is the prior of the latent variables and
$p(\bbx_n | \bbz_n, \bbbeta)$ is the likelihood model for generating
observations. For notation simplicity, we define $\bbtheta = (\bbalpha, \bbbeta)$.
Depending on the structure of $\bbz$, various DGMs have been developed, such as
the deep belief networks~\cite{Salakhutdinov:09,Lee:09}, deep sigmoid networks~\cite{Mnih:icml2014},
deep latent Gaussian models~\cite{danilo14icml}, and deep autoregressive models~\cite{Gregor:14}.
In this paper, we focus on the directed DGMs, which can be easily sampled from via an ancestral sampler.

However, in most cases learning DGMs is challenging due to the
intractability of posterior inference. The state-of-the-art
methods resort to stochastic variational methods under the maximum likelihood estimation (MLE) framework, $\hat \bbtheta = \argmax_{\bbtheta} \log p(\bbX | \bbtheta)$ (See the related work for alternative learning methods). Specifically,
let $q(\bbZ)$ be the variational distribution that approximates the true posterior $p(\bbZ | \bbX, \bbtheta)$.
A variational upper bound of the per sample negative log-likelihood (NLL) $-\log p(\bbx_n | \bbalpha, \bbbeta)$ is:
\setlength{\arraycolsep}{0.0em}
\begin{eqnarray}
\mathcal{L} (\bbtheta, q(\bbz_n ); \bbx_n) & \triangleq & \KL (q(\bbz_n) || p(\bbz_n | \bbalpha)) \nonumber \\ && -  \mathbb{E}_{q(\bbz_n)}[\log p(\bbx_n | \bbz_n, \bbbeta)] \nonumber,
\end{eqnarray}
where $\KL (q  || p )$ is the Kullback-Leibler (KL) divergence between distributions $q$ and $p$.
Then, $\mathcal{L} (\bbtheta, q(\bbZ); \bbX) \! \triangleq \! \sum_{n} \! \mathcal{L} (\bbtheta, q(\bbz_n) ; \bbx_n)$ upper bounds the full negative log-likelihood $-\log p(\bbX | \bbtheta)$.

It is important to notice that if we do not make restricting assumption on
the variational distribution $q$, the bound is tight by simply setting
$q(\bbZ) = p(\bbZ | \bbX, \bbtheta)$. That is, the MLE is equivalent to solving the variational
problem:
$\min_{\bbtheta, q(\bbZ)} \mathcal{L} (\bbtheta, q(\bbZ); \bbX)$.
However, since the true posterior is intractable except a handful of special cases, we must resort to approximation methods.
One common assumption is that the variational distribution is of some parametric form, $q_{\bbphi}(\bbZ)$, and
then we optimize the variational bound w.r.t the variational parameters $\bbphi$.
For DGMs, another challenge arises that the variational bound
is often intractable to compute analytically. To address this challenge, the early work
further bounds the intractable parts with tractable ones by introducing more
variational parameters~\cite{saul1996}. However, this technique increases the gap between
the bound being optimized and the log-likelihood, potentially resulting in
poorer estimates. Much recent progress~\cite{kingma14iclr,danilo14icml,Mnih:icml2014} has been made on hybrid Monte Carlo and variational methods, which
approximates the intractable expectations and their gradients over the parameters $(\bbtheta, \bbphi)$ via
some unbiased Monte Carlo estimates. Furthermore, to handle large-scale datasets,
stochastic optimization of the variational objective can be used with a suitable learning
rate annealing scheme. It is important to notice that variance reduction is a key part of these methods in
order to have fast and stable convergence.

Most work on directed DGMs has been focusing on the generative capability
on inferring the observations, such as filling in missing values~\cite{kingma14iclr,danilo14icml,Mnih:icml2014}, while
relatively insufficient work has been done on investigating the predictive power, except the recent advances~\cite{kingma14nips,springenberg16} for semi-supervised learning.
Below, we present max-margin deep generative models, which explore the
discriminative max-margin principle to improve the predictive
ability of the latent representations, while retaining the generative capability.

\subsection{Max-margin Deep Generative Models}

\begin{figure}[!t]
\centering
\includegraphics[width=.9\columnwidth]{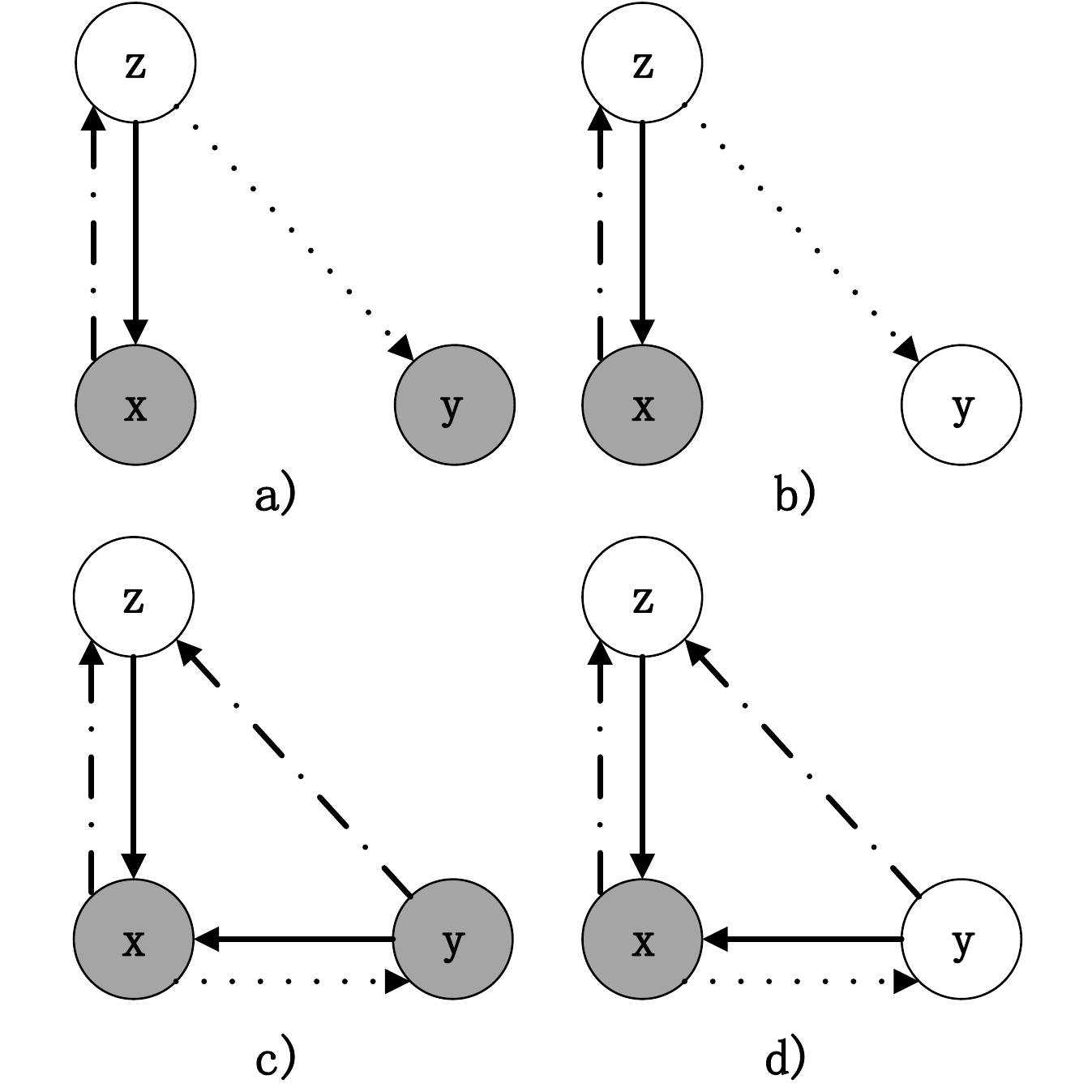}
\caption{a) and b): Graphical models of mmDGMs when labels are given or missing. c) and d): Graphical models of mmDCGMs when labels are given or missing. The solid line and the dash dot line represent the generative model and recognition model respectively. The dot line stands for the max-margin classifier. Compared with mmDGMs, mmDCGMs disentangle the label information from the latent variables and separate the pathways of inferring labels and latent variables.}
\label{fig:PGM}
\end{figure}

We first consider the fully supervised setting, where the training data
is a pair $(\bbx, y)$ with input features $\bbx \in \mathbb{R}^D$ and the groundtruth label $y$. Without loss of generality,
we consider the multiclass classification, where $y \in \mathcal{C} = \{1, \dots, M\}$.
As illustrated in Fig.~\ref{fig:PGM}, a max-margin deep generative model (mmDGM) consists of two components: (1) a deep generative model to
describe input features; and (2) a max-margin classifier to consider supervision. For the generative model, we
can in theory adopt any DGM that defines a joint distribution over $(\bbX, \bbZ)$ as in~Eq.~(\ref{eq:DGM-joint-dist}).
For the max-margin classifier, instead of fitting the input features into a conventional SVM,
we define the linear classifier on the latent representations, whose learning will be regularized
by the supervision signal as we shall see.
Specifically, if the latent representation $\bbz$ is given, we define the latent discriminant function $
F(y, \bbz, \bbeta; \bbx) = \bbeta^{\top} \bbf(y, \bbz),
$
where $\bbf(y, \bbz)$ is an $MK$-dimensional vector that concatenates $M$ subvectors, with the $y$th
being $\bbz$ and all others being zero, and $\bbeta$ is the corresponding weight vector.

We consider the case that $\bbeta$ is a random vector, following some prior distribution $p_0(\bbeta)$.
Then our goal is to infer the posterior distribution $p(\bbeta, \bbZ | \bbX, \bbY)$, which is typically
approximated by a variational distribution $q(\bbeta, \bbZ)$ for computational tractability. Notice that
this posterior is different from the one in the vanilla DGM.
We expect that the supervision information will bias the learned representations
to be more powerful on predicting the labels at testing.
To account for the uncertainty of $(\bbeta, \bbZ)$, we take the expectation and define the
discriminant function
$
F(y; \bbx) = \ep_{q}\left[ \bbeta^{\top} \bbf(y, \bbz) \right],
$
and the final prediction rule that maps inputs to outputs is:
\begin{eqnarray}\label{eq:mm-classifier}
\hat y  = \argmax_{y \in \mathcal{C}} F(y; \bbx ).
\end{eqnarray}
Note that different from the conditional DGM~\cite{kingma14nips}, which puts the class labels upstream and generates the latent representations as well as input data $\bbx$ by conditioning on $y$,
the above classifier is a downstream model in the sense that the supervision signal is
determined by conditioning on the latent representations.

\subsubsection{The Learning Problem}

We want to jointly learn the parameters $\bbtheta$ and infer the posterior distribution $q(\bbeta, \bbZ)$.
Based on the equivalent variational formulation of MLE, we define the joint learning problem as solving:
\begin{eqnarray}
\min_{\bbtheta, q(\bbeta, \bbZ), \bbxi} && \mathcal{L} (\bbtheta, q(\bbeta, \bbZ); \mathbf{X}) + C\sum_{n = 1}^N \xi_n \\
\forall n, y \in \mathcal{C},
\textrm{s.t. :} && \left\{ \begin{array}{ll}
\ep_q[\bbeta^{\top} \Delta \bbf_n(y)] \ge \Delta l_n(y) - \xi_n\\
\xi_n \ge 0,
\end{array} \right. \nonumber
\end{eqnarray}
where $\Delta \bbf_n(y) = \bbf(y_n, \bbz_n) - \bbf(y, \bbz_n)$ is the difference of the feature vectors;
$\Delta l_n(y)$ is the loss function that measures the cost to predict $y$ if the true label is $y_n$;
and $C$ is a nonnegative regularization parameter balancing the two components. 
In the objective, the variational bound is defined as
$\mathcal{L} (\bbtheta, q(\bbeta, \bbZ); \mathbf{X}) = \KL (q(\bbeta, \bbZ) || p_0(\bbeta, \bbZ | \bbalpha) )
 - \ep_q \left[ \log p(\bbX | \bbZ, \bbbeta)  \right] $, and the margin constraints are from the classifier~(\ref{eq:mm-classifier}).
If we ignore the constraints (e.g., setting $C$ at 0), the solution of $q(\bbeta, \bbZ)$ will be exactly the Bayesian posterior, and
the problem is equivalent to do MLE for $\bbtheta$.

By absorbing the slack variables, we can rewrite the problem in an unconstrained form:
\begin{equation}\label{eq:joint-problem}
 \min_{\bbtheta, q(\bbeta, \bbZ)} \mathcal{L} (\bbtheta, q(\bbeta, \bbZ); \mathbf{X}) + C \mathcal{R}(q(\bbeta, \bbZ; \mathbf{X})),
\end{equation}
where the hinge loss is: $
\mathcal{R}(q(\bbeta, \bbZ); \mathbf{X}) = \sum_{n = 1}^N\max_{y \in \mathcal{C}} (\Delta l_n(y) - \ep_q[\bbeta^{\top} \Delta \bbf_n(y)]).
$
Due to the convexity of $\max$ function, it is easy to verify that the hinge loss is an upper bound of the training error of classifier~(\ref{eq:mm-classifier}), that is,
$\mathcal{R}(q(\bbeta, \bbZ); \mathbf{X}) \geq \sum_n \Delta l_n( \hat{y}_n )$. Furthermore, the hinge loss is a convex functional over the variational distribution
because of the linearity of the expectation operator. These properties render the hinge loss as a good surrogate to optimize over.
Previous work has explored this idea to learn discriminative topic models~\cite{zhu12jmlr}, but with a restriction on the
shallow structure of hidden variables. Our work presents a significant extension to learn deep generative models, which pose
new challenges on the learning and inference.

\subsubsection{The Doubly Stochastic Subgradient Algorithm}\label{sec:algorithm}

The variational formulation of problem~(\ref{eq:joint-problem}) naturally suggests that
we can develop a variational algorithm to address the intractability
of the true posterior. We now present a new algorithm to solve problem~(\ref{eq:joint-problem}). Our method is a doubly stochastic generalization of the
Pegasos (i.e., Primal Estimated sub-GrAdient SOlver for SVM) algorithm~\cite{shai11pegasos} for the classic SVMs with fully observed input features,
with the new extension of dealing with a highly nontrivial structure of latent variables.

First,  we make the structured mean-field (SMF) assumption that
$q(\bbeta, \bbZ) = q(\bbeta) q_{\bbphi}(\bbZ)$. 
Under the assumption, we have the discriminant function
as
$\ep_q[\bbeta^{\top}  \Delta \bbf_n(y)] 
= \ep_{q(\bbeta)}[\bbeta^{\top}]  \ep_{q_{\bbphi}(\bbz^{(n)})}[\Delta \bbf_n(y)].$
Moreover, we can solve for the optimal solution of $q(\bbeta)$ in some analytical form.
In fact, by the calculus of variations, we can show that given the other parts the solution is
$
q(\bbeta) \propto p_0(\bbeta) \exp \Big( \bbeta^{\top} \sum_{n,y}  \omega_n^y \ep_{q_{\bbphi} }[ \Delta \bbf_n (y)] \Big),
$
where $\bbomega$ are the Lagrange multipliers (See~\cite{zhu12jmlr} for details).
If the prior is normal, $p_0(\bbeta) = \mathcal{N}(\bbzero, \sigma^2 \bbI)$, we have the normal posterior: $
q(\bbeta) = \mathcal{N}(\bblambda, \sigma^2 \bbI),~\textrm{where}~\bblambda = \sigma^2 \sum_{n,y} \omega_n^y  \ep_{q_{\bbphi} }[ \Delta \bbf_n(y) ].
$
Therefore, even though we did not make a parametric form assumption of $q(\bbeta)$, the above results show that the optimal
posterior distribution of $\bbeta$ is Gaussian.
Since we only use the expectation in the optimization problem and in prediction,
we can directly solve for the mean parameter $\bblambda$ instead of $q(\bbeta)$.
Further, in this case we can verify that $\KL(q(\bbeta) ||p_0(\bbeta)) =  \frac{||\bblambda||^2}{2\sigma^2}$
and then the equivalent objective function in terms of $\bblambda$ can be written as:
\begin{equation}\label{eq:learning_problem}
 \min_{\bbtheta, \bbphi, \bblambda} \mathcal{L} (\bbtheta, \bbphi; \mathbf{X}) + \frac{||\bblambda||^2}{2\sigma^2}
 +  C \mathcal{R}( \bblambda, \bbphi; \bbX),
\end{equation}
where $\mathcal{R}( \bblambda, \bbphi; \bbX) = \sum_{n = 1}^N \ell(\bblambda, \bbphi; \bbx_n) $ is the total hinge loss, and the per-sample hinge-loss is
$\ell(\bblambda, \bbphi; \bbx_n) = \max_{y \in \mathcal{C}} (\Delta l_n(y) - \bblambda^{\top}\ep_{q_{\bbphi}}[ \Delta \bbf_n(y)])$. Below, we present a doubly stochastic subgradient descent algorithm to solve this problem.

The {\it first stochasticity} arises from a stochastic estimate of the objective by random mini-batches.
Specifically, the batch learning needs to scan the full dataset to compute subgradients, which
is often too expensive to deal with large-scale datasets. One effective technique
is to do stochastic subgradient descent~\cite{shai11pegasos}, where at each iteration we randomly draw a mini-batch of
the training data and then do the variational updates over the small mini-batch.
Formally, given a mini batch of size $m$, we get an unbiased estimate of the objective:
\begin{equation}
\tilde{\mathcal{L}}_m := \frac{N}{m}\sum_{n = 1}^m \mathcal{L} (\bbtheta, \bbphi; \bbx_n) + \frac{||\bblambda||^2}{2\sigma^2}
  + \frac{NC}{m} \sum_{n = 1}^m \ell(\bblambda, \bbphi; \bbx_n). \nonumber
\end{equation}

The {\it second stochasticity} arises from a stochastic estimate of the per-sample variational bound and its subgradient,
whose intractability calls for another Monte Carlo estimator.

Formally, let $\bbz_n^l \sim q_{\bbphi}(\bbz | \bbx_n, y_n)$ be a set of samples from the variational distribution, where we explicitly put the conditions.
Then, the estimates of the per-sample variational bound and the per-sample hinge-loss are
\begin{equation}
\tilde{\mathcal{L}}(\bbtheta, \bbphi; \bbx_n) = \frac{1}{L}  \sum_l \log p(\bbx_n, \bbz_n^l | \bbbeta) -  \log q_{\bbphi}(\bbz_n^l) \nonumber
\end{equation}
and
\begin{equation}
\tilde{\ell}(\bblambda, \bbphi; \bbx_n) = \max_y  \Big(  \Delta l_n(y) - \frac{1}{L}   \sum_l  \bblambda^\top  \Delta \bbf_n(y, \bbz_n^l)  \Big)  \nonumber
\end{equation}
respectively,
where $\Delta \bbf_n(y, \bbz_n^l) = \bbf(y_n, \bbz_n^l) - \bbf(y, \bbz_n^l)$.
Note that $\tilde{\mathcal{L}}$ is an unbiased estimate of $\mathcal{L}$, while $\tilde{\ell}$ is a biased estimate of $\ell$.
Nevertheless, we can still show that $\tilde{\ell}$ is an upper bound estimate of $\ell$ under expectation.
Furthermore, this biasedness does not affect our estimate of the gradient.
In fact, by using the equality $\nabla_{\bbphi}q_{\bbphi}(\bbz) = q_{\bbphi}(\bbz) \nabla_{\bbphi} \log q_{\bbphi}(\bbz) $,
we can construct an unbiased Monte Carlo estimate of $\nabla_{\bbphi} (\mathcal{L}(\bbtheta, \bbphi; \bbx_n) + \ell(\bblambda, \bbphi; \bbx_n))$ as:
\setlength{\arraycolsep}{0.0em}\begin{eqnarray}\label{eq:var-grad}
\bbg_{\bbphi} &=& \frac{1}{L} \sum_{l=1}^L  \Big( \log p(\bbz_n^l, \bbx_n) - \log q_{\bbphi}(\bbz_n^l) \nonumber
 + C \bblambda^\top \Delta \bbf_n( \ytilde_n, \bbz_n^l ) \Big) \\
 &&\nabla_{\bbphi} \log q_{\bbphi}(\bbz_n^l) ,
\end{eqnarray}
where the last term roots from the hinge loss with the loss-augmented prediction
$\ytilde_n = \argmax_y (\Delta l_n(y) + \frac{1}{L} \sum_l \bblambda^\top \bbf(y, \bbz_n^l) )$.
For $\bbtheta$ and $\bblambda$, the estimates of the gradient $\nabla_{\bbtheta} \mathcal{L}(\bbtheta, \bbphi; \bbx_n)$ and the subgradient $\nabla_{\bblambda} \ell(\bblambda, \bbphi; \bbx_n)$ are easier, which are:
\begin{equation}
\bbg_{\bbtheta} = \frac{1}{L} \sum_l \nabla_{\bbtheta} \log p(\bbx_n, \bbz_n^l | \bbtheta), \nonumber
\end{equation}
and
\begin{equation}
\bbg_{\bblambda} = \frac{1}{L} \sum_l \left( \bbf(\ytilde_n, \bbz_n^l) - \bbf(y_n, \bbz_n^l) \right).  \nonumber
\end{equation}
Notice that the sampling and the gradient $\nabla_{\bbphi} \log q_{\bbphi}(\bbz_n^l)$
only depend on the variational distribution, not the underlying model.

\begin{algorithm}[t]
\caption{Doubly Stochastic Subgradient Algorithm}\label{alg:double-stochastic}
\begin{algorithmic}
   \STATE Initialize $\bbtheta$, $\bblambda$, and $\bbphi$
   \REPEAT
   \STATE draw a random mini-batch of $m$ data points
   \STATE draw random samples from noise distribution $p(\bbepsilon)$
   \STATE compute subgradient $\bbg = \nabla_{\bbtheta, \bblambda, \bbphi} \tilde{\mathcal{L}}(\bbtheta, \bblambda, \bbphi; \bbX^m, \bbepsilon)$ 
   \STATE update parameters $(\bbtheta, \bblambda, \bbphi)$ using subgradient $\bbg$.
   \UNTIL{Converge}
   \STATE {\bf return} $\bbtheta$, $\bblambda$, and $\bbphi$
\end{algorithmic}
\end{algorithm}

The above estimates consider the general case where the variational bound is intractable. In some cases,
we can compute the KL-divergence term analytically, e.g., when the prior and the variational distribution
are both Gaussian. 
In such cases, we 
only need to estimate the rest intractable part by sampling, which often reduces the variance~\cite{kingma14iclr}.
Similarly,
we could use the expectation of the features directly,
if it can be computed analytically,
in the computation of subgradients (e.g., $\bbg_{\bbtheta}$ and $\bbg_{\bblambda}$) instead of sampling, which again can lead to variance reduction.

With the above estimates of subgradients, we can use stochastic optimization methods such as SGD~\cite{shai11pegasos} and AdaM~\cite{kingma:15} to update the parameters, as outlined in Alg.~\ref{alg:double-stochastic}. Overall, our algorithm is a doubly stochastic generalization of Pegasos to deal with the highly nontrivial latent variables.

Now, the remaining question is how to define an appropriate variational distribution $q_{\bbphi}(\bbz)$ to obtain a robust estimate of the subgradients as well as the objective.
Two types of methods have been developed for unsupervised DGMs, namely, variance reduction~\cite{Mnih:icml2014} and auto-encoding variational Bayes (AVB)~\cite{kingma14iclr}.
Though both methods can be used for our models, we focus on the AVB approach.
For continuous variables $\bbZ$, under certain mild conditions  we can reparameterize the variational distribution $q_{\bbphi}(\bbz)$ using some simple variables $\bbepsilon$.
Specifically, we can draw samples $\bbepsilon$ from some simple distribution $p(\bbepsilon)$ and do the transformation $\bbz = \bbg_{\bbphi}(\bbepsilon, \bbx, y)$ to get the sample of the distribution $q(\bbz | \bbx, y)$. We refer the readers to~\cite{kingma14iclr} for more details.
In our experiments, we consider the special Gaussian case, where we assume that the variational distribution is a multivariate Gaussian with a diagonal covariance matrix:
\begin{eqnarray}\label{eq:recognition-model}
q_{\bbphi}(\bbz | \bbx, y) = \mathcal{N}( \bbmu(\bbx,y; \bbphi), \bbsigma^2(\bbx,y; \bbphi) ),
\end{eqnarray}
whose mean and variance are functions of the input data. This defines our recognition model.
Then, the reparameterization trick is as follows: we first draw standard normal variables $\bbepsilon^l \sim \mathcal{N}(0, \mathbf{I})$ and then do
the transformation $\bbz_n^l = \bbmu(\bbx_n,y_n;\bbphi) + \bbsigma(\bbx_n,y_n; \bbphi) \odot \bbepsilon^l$ to get a sample.
For simplicity, we assume that both the mean and variance are function of $\bbx$ only.
However, it is worth to emphasize that although the recognition model is unsupervised,
the parameters $\bbphi$ are learned in a supervised manner because the subgradient~(\ref{eq:var-grad}) depends on the hinge loss.
Further details of the experimental settings are presented in Sec.~\ref{sec:experiment_setting}.

\subsection{Conditional Variants for Semi-supervised Learning}

As collecting labeled data is often costly and time-consuming, semi-supervised learning (SSL)~\cite{zhu:book09} is an important setting, where the easy-to-get unlabeled data are leveraged to improve the quality. We now present an extension of mmDGMs to the semi-supervised learning scenario.

Given a labeled dataset $\mD_L = \{(\bbx_n, y_n )\}_{n=1}^{N_L}$ and an unlabeled dataset $\mD_U = \{\bbx_n\}_{n=1}^{N_U}$, where the size $N_U$ is typically much larger than $N_L$, the goal of SSL is to explore the intrinsic structures underlying the unlabeled data to help learn a classifier.
As the learning objective of mmDGMs consists of two parts---a data likelihood and a classification loss, a naive approach to considering unlabeled data is to simply ignore the loss term when the class label is missing. However, such ignorance leads to a weak coupling between the likelihood model and the classifier. Below, we present a conditional variant of mmDGMs, namely max-margin deep conditional generative models (mmDCGMs), to strongly couple the classifier and data likelihood.

Similar as in mmDGMs, an mmDCGM consists of two components: (1) a deep max-margin classifier to infer labels given data and (2) a class-conditional deep generative model to describe the joint distribution of the data, labels and latent variables. Fig.~\ref{fig:PGM} compares the graphical models of the mmDGM and mmDCGM.
Below, we present the learning objective of mmDCGM formally, which consists of several key components.
For notation simplicity, we will omit the parameters $\bbtheta$, $\bbphi$ and $\bblambda$ in the following formulae if no confusion arises.

{\bf Generative loss:} The first part of our learning objective is a generative loss to describe the observed data. For the labeled data $\bbx_n$ whose $y_n$ is visible, mmDCGM maximizes the joint likelihood for the pair $(\bbx_n, y_n)$, $\log p(\bbx_n, y_n)$, which is lower bounded by:
\begin{equation}
\mathcal{L}(\bbx_n,y_n) =  \ep_{q(\bbz_n|\bbx_n,y_n)}\left[ \log \frac{ p(\bbx_n| \bbz_n, y_n) p (\bbz_n) p(y_n)}{q(\bbz_n|\bbx_n,y_n)}\right].
\end{equation}
For the unlabeled data $\bbx_n$ whose $y_n$ is hidden, we can maximize the marginal likelihood $\log p(\bbx_n)$ by integrating out the hidden labels, whose variational lower-bound is:
\setlength{\arraycolsep}{0.0em}\begin{eqnarray}
\log p(\bbx_n) & \ge & \ep_{q(y | \bbx_n )} \ep_{q(\bbz_n | \bbx_n, y)}\left[ \log \frac{ p(\bbx_n| \bbz_n, y) p (\bbz_n) p (y)}{ q (y | \bbx_n) q(\bbz_n|\bbx_n, y)} \right] \nonumber \\
 & = & \ep_{q(y | \bbx_n )} \left[ \mathcal{L}(\bbx_n, y) \right] 
  + \mathcal{H}(q(y | \bbx_n)).
\end{eqnarray}
These lower-bounds were adopted in the previous method~\cite{kingma14nips}. However, one issue with this method is on the computational inefficiency when dealing with a large set of unlabeled data and a large number of classes. This is because we need to compute the lower-bounds of the joint likelihood for all possible $y \in
\mathcal{C}$ and for each unlabeled data point.

To make it computationally efficient, we propose to use the prediction of a classifier $\hat y_n = \arg \max \tilde q(y | \bbx_n)$ as a point estimation to approximate the full posterior $q(y | \bbx_n)$ to speed-up the inference procedure, where we denote the classifier by $\tilde q$ because it is not restricted to a specific form with a proper distribution over labels but is an unnormalized one trained under the max-margin principle. Indeed, the outputs of the classifier are real values transformed by linear operations, denoting the signed distance from the data to the hyperplanes defined by the weights. Consequently, the entropy term should be zero and the lower-bound turns out to be:
\begin{equation}\label{eqn:point_y}
\log p(\bbx_n) \ge
\ep_{q(\bbz_n | \bbx_n, \hat y_n)}\left[ \log \frac{ p(\bbx_n | \bbz_n, \hat  y_n) p (\bbz_n) p (\hat  y_n)}{ q(\bbz_n | \bbx_n, \hat  y_n)} \right].
\end{equation}
Note that the lower-bound is valid because we can view $\hat y_n = \arg \max \tilde q(y_n | \bbx_n)$ as a delta distribution.

With the above deviations, we define the overall generative loss as the summation of the negative variational bounds over $\mD_L$ and $\mD_U$:
\begin{eqnarray}\label{eqn:gene_loss}
-\mathcal{L_G}=&\sum_{(\bbx_n, y_n) \in \mD_L}& \ep_{q(\bbz_n | \bbx_n, y_n)}\left[ \log \frac{  p(\bbx_n, y_n, \bbz_n)}{q(\bbz_n | \bbx_n, y_n)} \right]  \nonumber \\
 & + \sum_{\bbx_n \in \mD_U} & \ep_{q(\bbz_n | \bbx_n, \hat y_n)}\left[ \log \frac{  p(\bbx_n, \hat y_n, \bbz_n)}{ q(\bbz_n|\bbx_n, \hat y_n)}\right].
\end{eqnarray}

{\bf Hinge loss:} The second part of our learning objective is a hinge loss on the labeled data. Specifically, though the labeled data can contribute to the training of the classifier $\tilde q(y|\bbx)$ implicitly through the objective function in Eqn.~(\ref{eqn:gene_loss}), it has been shown that adding a predictive loss for the labeled data can speed-up convergence and achieve better results~\cite{kingma14nips,maaloe16}. Here, we adopt the similar idea by introducing a hinge loss as the discriminative regularization for the labeled data:
\begin{equation}
\mathcal{L_L} =  \sum_{(\bbx_n, y_n) \in \mD_L} \max_{ y \in \mathcal{C}} (\Delta l_n (y) + \bblambda^{\top}\ep_{q_{\bbphi}}[ \Delta \bbf_n(y)]),
\end{equation}
which is the same as in the fully supervised case.

{\bf Hat loss:} The third part of our learning objective is a hat loss on the unlabeled data. Specifically, as $N_U$ is typically much larger than $N_L$ in the semi-supervised learning, it is desirable that the unlabeled data can regularize the behaviour of the classifier explicitly. To this end, we further propose a max-margin ``hat loss''~\cite{zhu:book09} for the unlabeled data as follows:
\begin{equation}
\mathcal{L_U} =  \sum_{\bbx_n \in \mD_U} \max_{ y \in \mathcal{C}} (\Delta l_{\hat y_n} (y) + \bblambda^{\top}\ep_{q_{\bbphi}}[ \Delta \bbf_{\hat y_n }(y)]),
\end{equation}
where $\Delta \bbf_{\hat y_n }(y) = \bbf(\hat y_n, \bbz_n) - \bbf(y, \bbz_n)$ and
$\Delta l_{\hat y_n }(y)$ is an function that indicates whether $y$ equals to the prediction $\hat y_n$ or not. Namely, we treat the prediction $\hat y_n$ as putative label and apply the hinge loss function on the unlabeled data. This function is called the hat loss due to its shape in a binary classification example~\cite{zhu:book09}. Intuitively,
the hinge loss enforces the predictor to make prediction correctly and confidently with a large margin for labeled data, while the hat loss only requires the predictor to make decision confidently for unlabeled data.
The hat loss, which has been originally proposed in S3VMs~\cite{vapnik:book}, assumes that decision boundary tends to lie on low-density areas of the feature space. In such shallow models, the correctness of the assumption heavily depends on the true data distribution, which is fixed but unknown. However, the constraint is much more relaxed when building upon the latent feature space learned by a deep model as described in our method. In practice, the predictive performance of mmDCGMs is improved substantially by adding this regularization, which will be shown in Sec.~\ref{sec:ssl-results}.

{\bf Label-balance regularization:} The last part of our learning objective is a regularization term to balance the possible label predictions on the unlabeled data. Specifically, one practical problem of semi-supervised learning is the imbalance of the predictions~\cite{zhu:book09}, that is, a classifier may classify most of the unlabeled points as a same class. To address this problem, we introduce a balance constraint for multiclass semi-supervised learning:
\begin{equation}
 \forall y \in \mathcal{C}, \frac{1}{N_U} \sum_{\bbx_n \in \mD_U} \Delta l_{\hat y_n} (y) = \frac{1}{N_L} \sum_{(\bbx_n, y_n) \in \mD_L} \Delta l_{n} (y),
 \label{eqn:constraint_hard}
\end{equation}
which assumes that the distribution of the predictions of unlabeled data should be the same as that of the groundtruth labels in the labeled set. However, both sides in Eqn.~(\ref{eqn:constraint_hard}) are summations of indicator functions, which are non-differentiable with respect to $\bblambda$. Therefore, we cannot optimize $\bblambda$ based on gradient methods to satisfy this constraint directly. Here, we relax the constraint (\ref{eqn:constraint_hard})  as: $\forall y \in \mathcal{C},$
\begin{equation}
\label{eqn:constraint_relax}
\frac{1}{N_U} \sum_{\mD_U} \Delta l_{\hat y_n} (y) F(y; \bbx_n) = \frac{1}{N_L} \sum_{\mD_L} \Delta l_{n} (y) F(y; \bbx_n),
\end{equation}
where $F(y; \bbx_n) = \bblambda^{\top}\ep_{\tilde q}[ \bbf(y, \bbz)]$ and we simplify the summation notation.
Given certain class $y$, the left hand side selects the unlabeled data whose predictions equal to $y$ according to the indicator functions, and adds the corresponding activations (discriminant functions of $y$ divided by a factor $N_U$) together. The right hand side computes this normalized activations with indicator functions in same class for the labeled data. Note that $F(y; \bbx_n)$ is no smaller than $F(y'; \bbx_n)$ for any other $y'$ due to the definitions of the prediction $\hat y_n$ and the indicator function $\Delta l_{\hat y_n}(y)$.
The gradients in the relaxed version are still not well-defined due to the indicator functions.
However, assuming that the predictions $\hat y_n$ are given, both sides in Eqn.~(\ref{eqn:constraint_relax}) are summations without indicator functions, which are differentiable with respect to $\bblambda$.
In our experiments, we indeed ignore the dependency of the indicator functions on $\bblambda$ and approximate the total gradients by the gradients of the cumulative activations. This approximation does not work for the constraint in Eqn.~(\ref{eqn:constraint_hard}) because both sides turn out to be scalars given $\hat y_n$ and the gradient with respect to $\bblambda$ is zero almost everywhere, which cannot be used to optimize parameters.
In fact,
the relaxed constraint balances the predictions of unlabeled data according to the groundtruth implicitly, under the further assumption that the cumulative activation is proportional to the number of predictions for any $y$.
Intuitively, if the cumulative activation of the selected unlabeled data in certain class $y$ is larger than that of the labeled data, then probably the predictor classifies some unlabeled data as $y$ incorrectly. Consequently, the $\bblambda$ is updated to reduce the activations, and then the number of predictions in this class will decrease because $F(y; \bbx_n) $ may be smaller than $ F(y'; \bbx_n)$ for some other $y'$.
Moreover, as hard constraints are unlikely to satisfy in practice, we further relax them by using a regularization penalty in the common $L_2$-norm:
\begin{equation}
\mathcal{L_B} = \sqrt{\sum_{y \in \mathcal{C}}\left( \frac{\sum_{\mD_U} \Delta l_{\hat y_n} (y) F(y; \bbx)}{N_U} - \frac{\sum_{\mD_L} \Delta l_{n} (y) F(y; \bbx)}{N_L} \right)^2}. \nonumber
\end{equation}

With the above sub-objectives, our final objective function is a weighted sum:
\begin{equation}\label{eqn:final_obj}
\mathcal{L} = \mathcal{L_G} + \alpha(\mathcal{L_L} + \alpha_{\mU}\mathcal{L_U} + \alpha_{\mathcal{B}}\mathcal{L_B}),
\end{equation}
where $\alpha$, $\alpha_{\mU}$ and $\alpha_{\mathcal{B}}$ are hyper-parameters that control the relative weights for the corresponding terms. We will discuss the choice of each value in Sec.~\ref{sec:experiment_setting}.

\begin{figure*}
\centering
\includegraphics{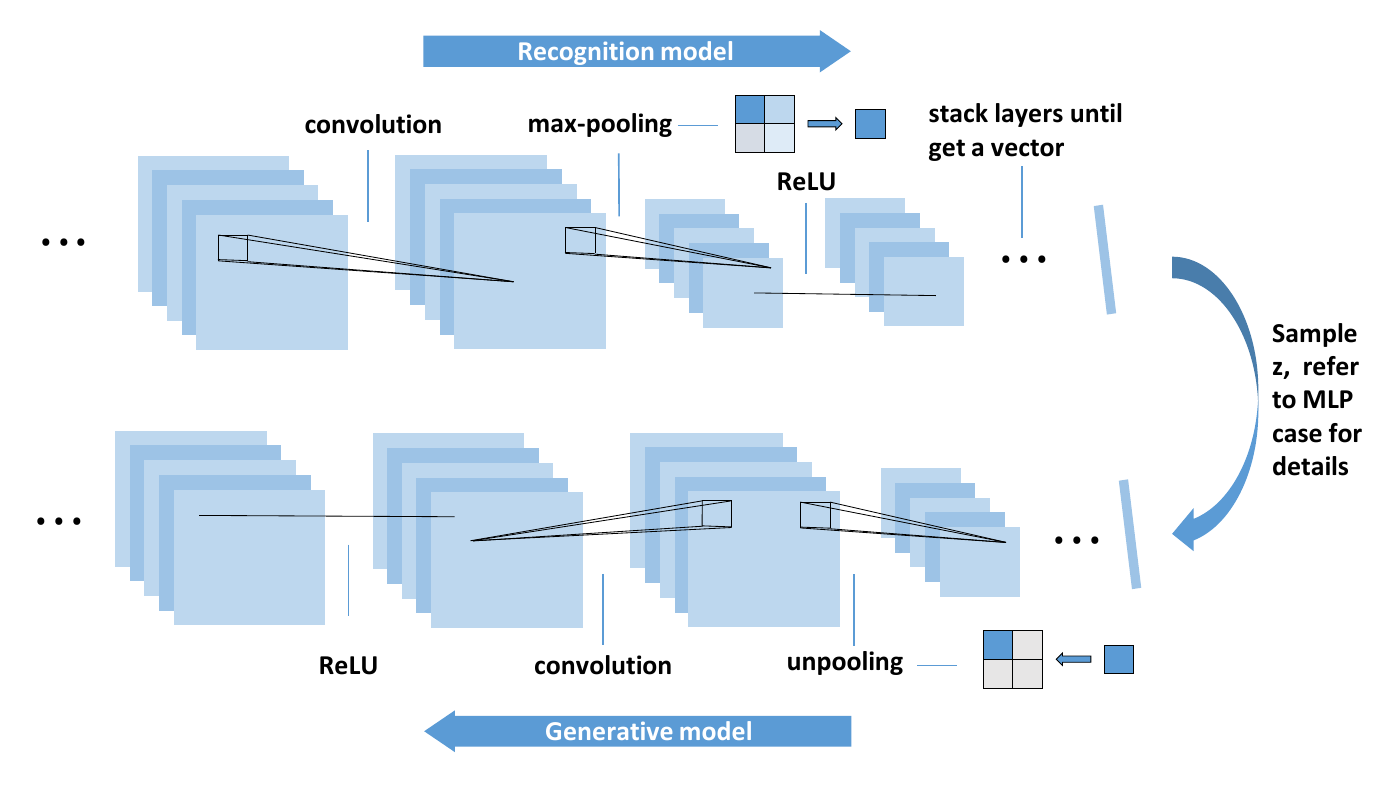}
\caption{Network architecture of Conv-MMVA with conv-net in the recognition model and unconv-net in the generative model (best view in color).}
\label{architecture-cmmva}
\end{figure*}

To optimize the overall learning objective, we still use our doubly stochastic algorithm described in Sec.~\ref{sec:algorithm} to compute the unbiased subgradient estimations for all of the parameters and perform updates.
Specifically, given a mini-batch of data consisting of labeled data $\mathcal{M_L} = \{(\bbx_n, y_n)\}_{n=1}^{m_L}$ and unlabeled data $\mathcal{M_U} = \{\bbx_n\}_{n=1}^{m_U}$,
we sequentially
\begin{enumerate}
\item predict $\hat y_n$ using the classifier for each $\bbx_n \in \mathcal{M_U}$;
\item plug in the predictions of the unlabeled data $\hat y_n$ and the groudtruth of the labeled data $y_n$ into the indicator functions in the label-balance regularization;
\item take (sub-)gradient with respect to all parameters in the generative model, recognition model and classfier to optimize the final objective~(\ref{eqn:final_obj});
\item approximate the (sub-)gradients with intractable expectations using the techniques described in Sec.~\ref{sec:algorithm} and update parameters.
\end{enumerate}
Though the objective in semi-supervised learning is complex, our method works well in practice.

\section{Experiments}

We now present the experimental results in both supervised and semi-supervised learning settings. Our results on several benchmark datasets demonstrate that both mmDGMs and mmDCGMs are highly competitive in classification while retaining the generative ability, under the comparison with various strong competitors.

\subsection{Experiment Settings}
\label{sec:experiment_setting}

Though mmDGMs and mmDCGMs are applicable to any DGMs that define a joint distribution of $(\bbX, \bbZ)$ and $(\bbX, \bbZ, \bbY)$ respectively, we concentrate on the Variational Auto-encoder (VA)~\cite{kingma14iclr} and Conditional VA~\cite{kingma14nips} in our experiments. We consider two types of recognition models: multiple layer perceptrons (MLPs) and convolutional neural networks (CNNs).
We denote our mmDGM with MLPs by {\bf MMVA}.
To perform classification using VA which is unsupervised, we first learn the feature representations by VA, and then build a linear SVM classifier on
these features using the Pegasos stochastic subgradient algorithm~\cite{shai11pegasos}. This baseline will be denoted by {\bf VA+Pegasos}.
The corresponding models with CNNs are denoted by {\bf Conv-MMVA} and {\bf Conv-VA+Pegasos} respectively. We denote our mmDCGM with CNNs by {\bf Conv-MMCVA}. We implement all experiments based on Theano~\cite{Bastien-Theano-2012}.~
\footnote{Source code and more detailed settings can be found at https://github.com/thu-ml/mmdcgm-ssl.}

\subsubsection{Datasets and Preprocessing}

We evaluate our models on the widely adopted MNIST~\cite{Lecun:98}, SVHN~\cite{Netzer:11} and small NORB~\cite{lecun2004learning} datasets. MNIST consists of handwritten digits of 10 different classes (0 to 9). There are 50,000 training samples, 10,000 validating samples and 10,000 testing samples and each one is of size $28 \times 28$. SVHN is a large dataset consisting of color images of size $32 \times 32$. The task is to recognize the center digits in natural scene images. We follow the work~\cite{Sermanet:12,goodfellow:13} to split the dataset into 598,388 training data, 6,000 validating data and 26,032 testing data.
The small NORB dataset consisits of gray images distributed across 5 general classes: animal, human, airplane, truck and car. Both the training set and testing set in NORB contain 24,300 samples with different lighting conditions and azimuths. We  down-sample the images to size of $32 \times 32$ as in~\cite{maaloe16} and split 1,000 samples from the training set as the validating data if required.

For fair comparison in supervised learning on SVHN, we perform Local Contrast Normalization (LCN) in the experiment of the Conv-MMVA following ~\cite{Sermanet:12,goodfellow:13} and set the distribution of $\bbx$ given $\bbz$ as Gaussian. In other cases, we just normalize the data by a factor of 256 and choose Bernoulli as the distribution of data.

\subsubsection{Supervised Learning}

In mmDGMs, the recognition network and the classifier share layers in computation.
The mean and variance of the latent variable $\bbz$ are transformed from the last layer of the recognition model through an affine transformation.
It should be noticed that we could use not only the expectation of $\bbz$ but also the activation of any layer in the recognition model as features.
The only theoretical difference is from where we add a hinge loss regularization to the gradient and back-propagate it to previous layers.
In all of the experiments, the mean of $\bbz$ has the same nonlinearity but typically much lower dimension than the activation of the last layer in the recognition model,
and hence often leads to a worse performance. We use different features in MMVA and Conv-MMVA, which will be explained below. We use AdaM~\cite{kingma:15} to optimize parameters in all of the models.
Although it is an adaptive gradient-based optimization method,
we decay the global learning rate by a factor after sufficient number of epochs to ensure a stable convergence.

In MMVA, we follow the settings in~\cite{kingma14nips} to compare both generative and discriminative capacity of VA and MMVA. Both the recognition and generative models employ a two-layer MLP with 500 hidden units in each layer and the dimension of the latent variables is 50. We choose $C = 15$ as default in MMVA. We concatenate the activations of 2 layers as the features used in the supervised tasks. We illustrate the network architecture of MMVA in Appendix A.

In Conv-MMVA, we use standard CNNs~\cite{Lecun:98} with convolution and max-pooling operation as the recognition model to obtain more competitive classification results.
For the generative model, we use unconvnets~\cite{Dosovitskiy:2014} with a ``symmetric'' structure as the recognition model, to reconstruct the input images approximately.
More specifically, the top-down generative model has the same structure as the bottom-up recognition model but replacing max-pooling with unpooling operation~\cite{Dosovitskiy:2014}
and applies unpooling, convolution and rectification in order. Typically, there are 5 or 6 convolutional layers in the generative model and the recognition model and the kernel size is either 5 or 3, depending on the data. The total number of parameters is comparable with previous work~\cite{goodfellow:13,Lin:14,Lee:15} and the split of the training sets is the same. For simplicity, we do not involve mlpconv layers~\cite{Lin:14,Lee:15} and contrast normalization layers in our recognition model,
but they are not exclusive to our model. We set $C=10^3$ on MNIST and $C=10^4$ on SVHN as default.  We use the activations of the last deterministic layer as the features. We illustrate the network architecture of Conv-MMVA with Gaussian hidden variables and Bernoulli visible variables in Fig.~\ref{architecture-cmmva}.

\subsubsection{Semi-supervised Learning}

The mmDCGM separates the classifier and the recognition model of the latent variables completely, which allows us to simply combine the state-of-the-art classifier and deep generative models together without competition. We only consider the convolutional neural networks here and adopt advanced techniques including global average pooling~\cite{lin2014network} and batch normalization~\cite{ioffe2015batch} to boost the performance of our Conv-MMCVA. The architecture of the max-margin classifier refers to that of the discriminator in~\cite{springenberg16} and the generative model is similar with the Conv-MMVA but concatenates the feature maps and additional label maps in one-hot encoding format at each layer as in~\cite{radford2015unsupervised}. Similar with Conv-MMVA, the depth of each convolutional networks is 5 or 6.
We set $\alpha = 0.1$ according to the conditional VAE~\cite{kingma14nips}. We optimize $\alpha_{\mU}$ and $\alpha_{\mathcal{B}}$ with a search grid $\{..., 0.01, 0.03, 0.1, 0.3, 1, 3 ...\}$ in terms of the validation classification error of a shallow S3VM on MNIST given 100 labels. The best values are $\alpha_{\mU} = 3$ and $\alpha_{\mathcal{B}} = 0.001$ and we fix them in our Conv-MMCVA across all of the datasets. Other hyper-parameters including the anneal strategy and batch size are chosen according to the validation generative loss. Once the hyperparameters are fixed, we run our model for 10 times with different random splits of the labeled and unlabeled data, and we report the mean and the standard deviation of the error rates.

\begin{table}[!t]
\renewcommand{\arraystretch}{1.3}
\caption{Error rates (\%) on the MNIST dataset given full labeled data.}
\centering
\begin{tabular}{lc}
\hline
Model & Error Rate \\
\hline
{\it VA+Pegasos} & 1.04 \\
{\it VA+Class-conditionVA}~\cite{kingma14nips} & 0.96 \\
{\it MMVA} &  0.90 \\
\hline
{\it Conv-VA+Pegasos} & 1.35\\
{\it Conv-MMVA} & 0.45\\
\hline
{\it Stochastic Pooling}~\cite{Zeiler:13} & 0.47\\
{\it Network in Network}~\cite{Lin:14} & 0.47\\
{\it Maxout Network}~\cite{goodfellow:13} & 0.45\\
{\it DSN}~\cite{Lee:15} & 0.39\\
\hline
\label{mnist-basic-table}
\end{tabular}
\end{table}

\begin{table}[!t]
\renewcommand{\arraystretch}{1.3}
\caption{Error rates (\%) on the SVHN dataset given full labeled data.}
\centering
\begin{tabular}{lc}
\hline
Model & Error Rate \\
\hline
{\it Conv-VA+Pegasos} & 25.3 \\
{\it Conv-MMVA} & 3.09\\
\hline
{\it CNN}~\cite{Sermanet:12} & 4.9 \\
{\it Stochastic Pooling}~\cite{Zeiler:13} & 2.80\\
{\it Maxout Network}~\cite{goodfellow:13} & 2.47\\
{\it Network in Network}~\cite{Lin:14} & 2.35\\
{\it DSN}~\cite{Lee:15} & 1.92\\
\hline
\label{svhn-basic-table}
\end{tabular}
\end{table}

\subsection{Results with Supervised Learning}
\label{sec:results}

We first present the results in the supervised learning setting. Specifically, we evaluate the predictive and generative performance of our MMVA and Conv-MMVA on the MNIST and SVHN datasets in various tasks, including classification, sample generation, and missing data imputation.

\begin{figure*}[!t]
\centering
\subfigure[VA ]{\includegraphics[width=0.49\columnwidth]{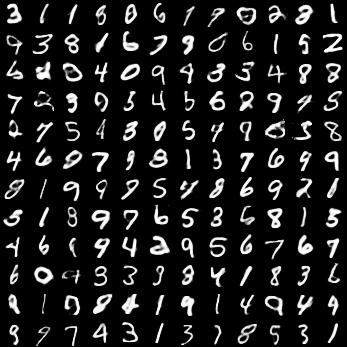}}
\subfigure[MMVA]{\includegraphics[width=0.49\columnwidth]{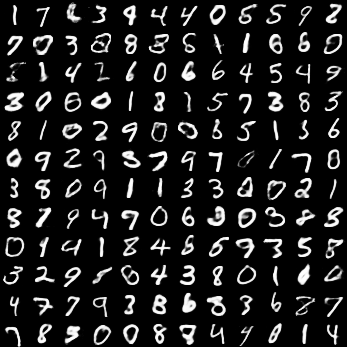}}
\subfigure[Conv-VA]{\includegraphics[width=0.49\columnwidth]{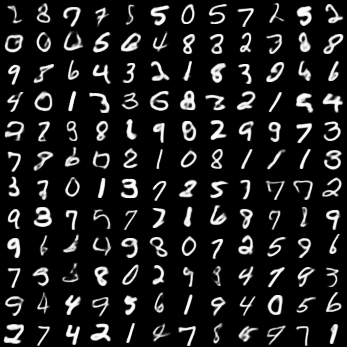}}
\subfigure[Conv-MMVA]{\includegraphics[width=0.49\columnwidth]{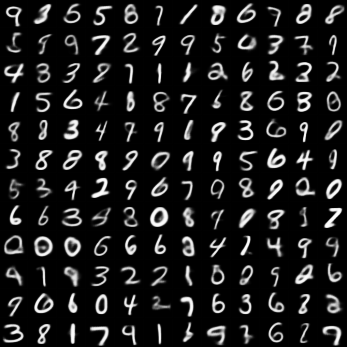}}
\caption{Generation on MNIST. (a-b): images randomly generated by VA and MMVA respectively;
(c-d): images randomly generated by Conv-VA and Conv-MMVA respectively. Our mmDGMs retain similar ability as the baselines to generate images.}
\label{va_sample}
\end{figure*}

\begin{figure*}[!t]
\centering
\subfigure[Training data]{\includegraphics[width=0.49\columnwidth]{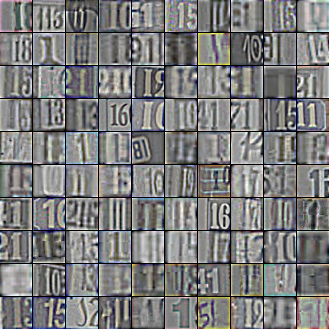}}
\subfigure[Conv-VA]{\includegraphics[width=0.49\columnwidth]{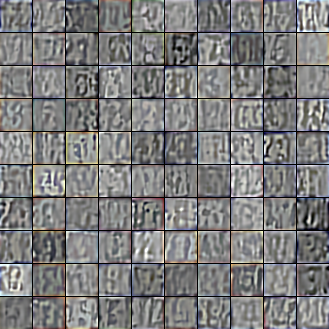}}
\subfigure[Conv-MMVA ($C=10^3$)]{\includegraphics[width=0.49\columnwidth]{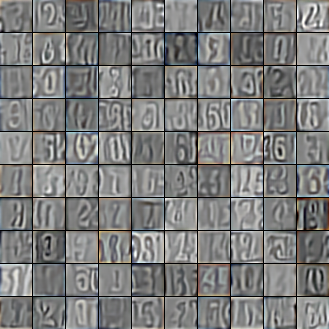}}
\subfigure[Conv-MMVA ($C=10^4$)]{\includegraphics[width=0.49\columnwidth]{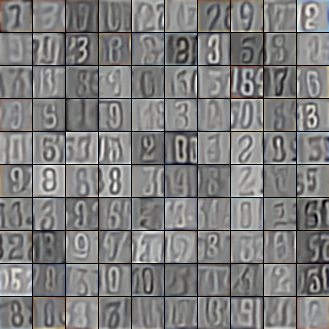}}
\caption{Generation on SVHN.
(a): training data preprocessed by LCN;
(b): samples randomly generated by Conv-VA;
(c-d): samples randomly generated by Conv-MMVA when $C=10^3$ and $C=10^4$ respectively.}
\label{svhn_sample}
\end{figure*}

\subsubsection{Predictive Performance}

\begin{table}[!t]
\renewcommand{\arraystretch}{1.3}
\caption{Effects of $C$ on the MNIST dataset in Conv-MMVA.}
\centering
\begin{tabular}{lcc}
\hline
C & Error Rate (\%) & Lower Bound \\
\hline
0 & 1.35      &  -93.17  \\
1 &   1.86   &  -95.86 \\
$10$ &  0.88   &   -95.90 \\
$10^{2}$ &  0.54    & -96.35  \\
$ 10^{3}$ &  0.45   &  -99.62 \\
$ 10^{4}$ &  0.43 &  -112.12  \\
\hline
\label{effect-c}
\end{tabular}
\end{table}

We test both MMVA and Conv-MMVA on the MNIST dataset. In MLP case, the first three rows in Table~\ref{mnist-basic-table} compare {VA+Pegasos}, {VA+Class-condtionVA} and {MMVA},
where {VA+Class-condtionVA} refers to the best fully supervised model in~\cite{kingma14nips}.
Our model outperforms the baselines significantly.
We further use the t-SNE algorithm~\cite{Maaten:08} to embed the features learned by VA and MMVA on 2D plane, which again demonstrates the stronger discriminative ability of MMVA (See Appendix B for details).

In CNN cases, Table~\ref{effect-c} shows the effect of $C$ on classification error rate and variational lower bound.
Typically, as $C$ gets lager, Conv-MMVA learns more discriminative features and leads to a worse estimation of data likelihood.
However,
if $C$ is too small,
the supervision is not enough to lead to predictive features.
Nevertheless, $C = 10^3$ is quite a good trade-off between the classification performance and generative performance.
In this setting, the classification performance of our Conv-MMVA model is comparable to the state-of-the-art fully discriminative networks with comparable architectures and number of parameters, shown in the last four rows of Table~\ref{mnist-basic-table}.

We focus on Conv-MMVA on the SVHN datset as it is more challenging. Table~\ref{svhn-basic-table} shows the predictive performance on SVHN. In this harder problem, we observe a larger improvement by Conv-MMVA as compared to Conv-VA+Pegasos, suggesting
that DGMs benefit a lot from max-margin learning on image classification.
We also compare Conv-MMVA with state-of-the-art results.
To the best of our knowledge, there is no competitive generative models to classify digits on the SVHN dataset with full labels. 

\subsubsection{Generative Performance}

We investigate the generative capability of MMVA and Conv-MMVA on generating samples.
Fig.~\ref{va_sample} and Fig.~\ref{svhn_sample} illustrate the images randomly sampled from VA and MMVA models on MNIST and SVHN respectively,
where we output the expectation of the value at each pixel to get a smooth visualization. Fig.~\ref{svhn_sample} demonstrates the benefits from jointly training of DGMs and max-margin classifiers.
Though Conv-VA gives a tighter lower bound of data likelihood and reconstructs data more elaborately,
it fails to learn the pattern of digits in a complex scenario and could not generate meaningful images.
In this scenario, the hinge loss regularization on recognition model is useful for generating main objects to be classified in images.

\begin{table}[!t]
\renewcommand{\arraystretch}{1.2}
\caption{MSE on MNIST data with missing values in the testing procedure.}
\centering
\begin{tabular}{lcccc}
\hline
Noise Type & VA & MMVA & Conv-VA & Conv-MMVA \\
\hline
Rand-Drop (0.2) & \textbf{0.0109} & 0.0110 & 0.0111  & 0.0147 \\
Rand-Drop (0.4) & \textbf{0.0127} & \textbf{0.0127} & \textbf{0.0127}  & 0.0161 \\
Rand-Drop (0.6) & 0.0168 & \textbf{0.0165} & 0.0175  & 0.0203 \\
Rand-Drop (0.8) & 0.0379 & \textbf{0.0358} & 0.0453  & 0.0449 \\
\hline
Rect (6 $\times$ 6) & 0.0637 & 0.0645 & \textbf{0.0585}  & 0.0597  \\
Rect (8 $\times$ 8) & 0.0850 & 0.0841 & 0.0754  & \textbf{0.0724}  \\
Rect (10 $\times$ 10) & 0.1100 & 0.1079 &  0.0978  & \textbf{0.0884} \\
Rect (12 $\times$ 12) & 0.1450 & 0.1342 &  0.1299  & \textbf{0.1090} \\
\hline
\label{deniose-mse}
\end{tabular}
\end{table}

\subsubsection{Missing Data Imputation and Classification}

\begin{figure*}[!t]
\centering
\subfigure[Rand-Drop (0.6)]{\includegraphics[width=.95\columnwidth]{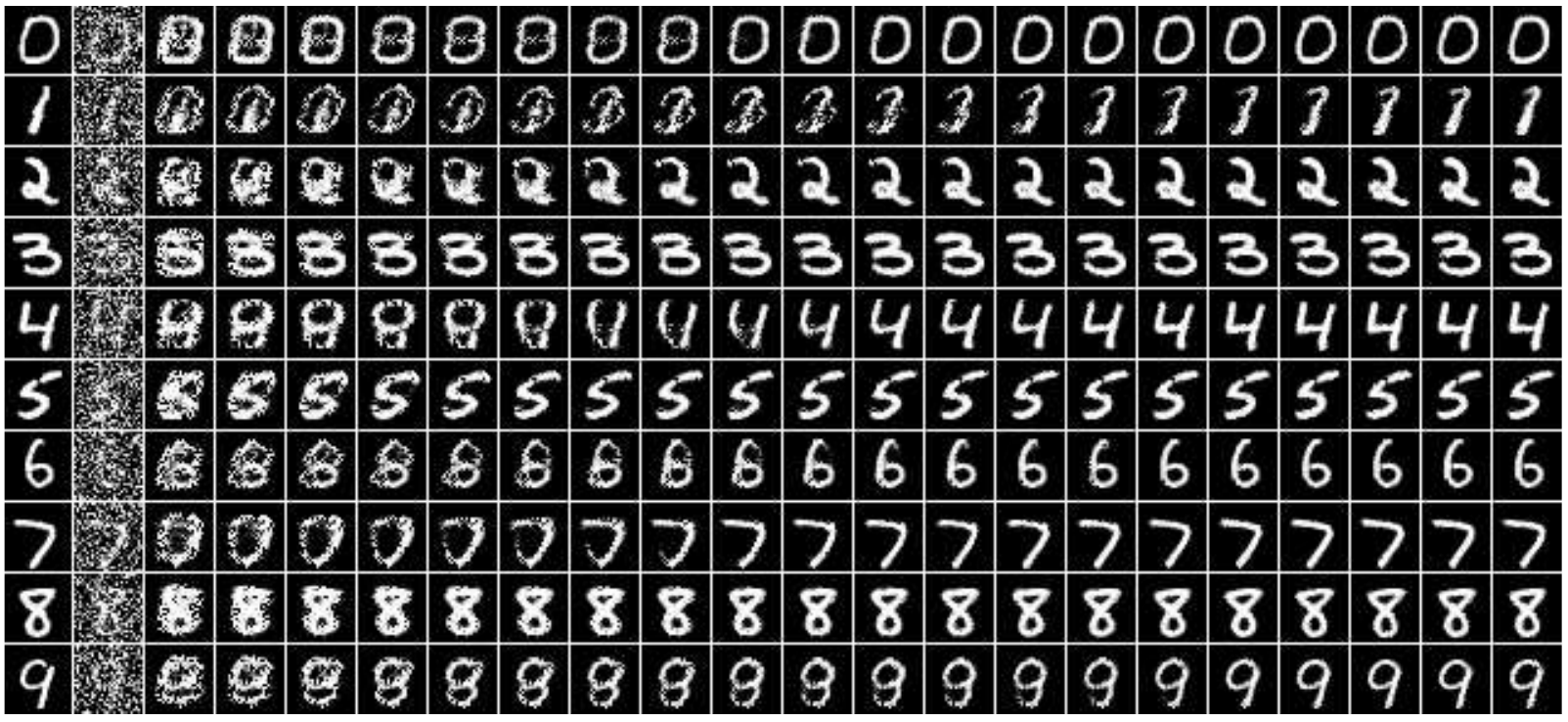}}
\subfigure[Rect ($12 \times 12$)]{\includegraphics[width=.95\columnwidth]{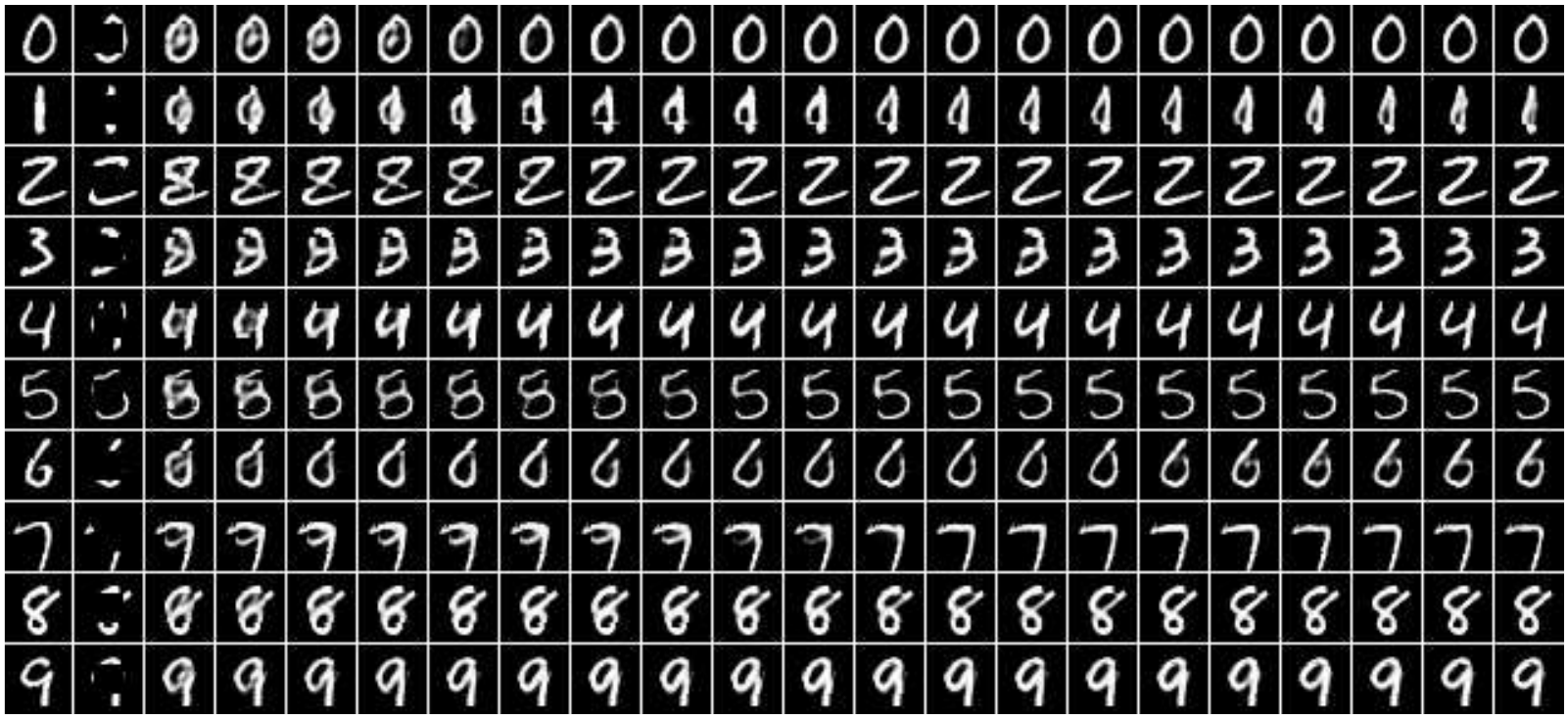}}
\caption{Imputation results of MMVA in two noising conditions: column 1 shows the true data; column 2 shows the perturbed data;
and the remaining columns show the imputations for 20 iterations.}
\label{denoise}
\end{figure*}

We further test MMVA and Conv-MMVA on the task of missing data imputation.
For MNIST, we consider two types of missing values~\cite{little87missing}:
(1) {\bf Rand-Drop}: each pixel is missing randomly with a pre-fixed probability;
and (2) {\bf Rect}: a rectangle located at the center of the image is missing.
Given the perturbed images,
we uniformly initialize the missing values between 0 and 1,
and then iteratively do the following steps:
(1) using the recognition model to sample the hidden variables;
(2) predicting the missing values to generate images; and (3) using the refined images as the input of the next round.
For SVHN,
we do the same procedure as in MNIST but initialize the missing values with Guassian random variables as the input distribution changes.

Intuitively, generative models with CNNs could be more powerful on learning patterns and high-level structures, while generative models with MLPs learn more to
reconstruct the pixels in detail.
This conforms to the MSE results shown in Table~\ref{deniose-mse}: Conv-VA and Conv-MMVA outperform VA and MMVA with a missing rectangle,
while VA and MMVA outperform Conv-VA and Conv-MMVA with random missing values.
Compared with the baselines, mmDGMs also make more accurate completion when large patches are missing. All of the models infer missing values for 100 iterations.

We visualize the inference procedure of MMVA in Fig.~\ref{denoise}.
Considering both types of missing values, MMVA could infer the unknown values and refine the images in several iterations even with a large ratio of missing pixels. More visualization results on MNIST and SVHN are presented in Appendix C.

\begin{table}[!t]
\renewcommand{\arraystretch}{1.2}
\caption{Error rates(\%) with missing values on MNIST.}
\centering
\begin{tabular}{lccc}
\hline
Noise Level &  CNN & Conv-VA & Conv-MMVA\\
\hline
Rect (6 $\times$ 6) & 7.5 & 2.5 & \textbf{1.9} \\
Rect (8 $\times$ 8) & 18.8 & 4.2 & \textbf{3.7}  \\
Rect (10 $\times$ 10) & 30.3 & 8.4 &  \textbf{7.7} \\
Rect (12 $\times$ 12) & 47.2 & 18.3 &  \textbf{15.9} \\
\hline
\label{errorwithmissing}
\end{tabular}
\end{table}

We further present classification results with missing values on MNIST in Table~\ref{errorwithmissing}. CNN makes prediction on the incomplete data directly. Conv-VA and Conv-MMVA infer missing data for 100 iterations at first and then make prediction on the refined data. In this scenario, Conv-MMVA outperforms both Conv-VA and CNN, which demonstrates the advantages of our mmDGMs, which have both strong discriminative and generative capabilities.

Overall, mmDGMs have comparable capability of inferring missing values
and prefer to learn high-level patterns instead of local details.

\subsection{Results with Semi-supervised Learning}
\label{sec:ssl-results}

We now present the predictive and generative results on MNIST, SVHN and small NORB datasets given partially labeled data.

\subsubsection{Predictive Performance}

\begin{table}[!t]
\renewcommand{\arraystretch}{1.2}
\caption{Error rates (\%) on (partially) labeled MNIST dataset.}
\centering
\begin{tabular}{lccc}
\hline
Algorithm & $n=100$ & $n=1000$  & ALL\\
\hline
{\it M1+M2}~\cite{kingma14nips} & 3.33 ($\pm0.14$) & 2.4 ($\pm0.02$)  & 0.96\\
{\it VAT}~\cite{miyato2015distributional} & 2.33 & 1.36 &  0.64 \\
{\it Ladder}~\cite{rasmus15} & 1.06 ($\pm 0.37$) & \textbf{0.84} ($\pm 0.08$)  & 0.57  \\
{\it CatGAN}~\cite{springenberg16} & 1.91 ($\pm 0.10$) & 1.73 ($\pm 0.18$)& 0.91 \\
{\it ADGM}~\cite{maaloe16} & 0.96 ($\pm 0.02$) & - & - \\
{\it SDGM}~\cite{maaloe16} & 1.32 ($\pm 0.07$) & - & -  \\
\hline
{\it Conv-CatGAN}~\cite{springenberg16} & 1.39 ($\pm 0.28$) & -& \textbf{0.48}\\
{\it Improved-GAN}~\cite{salimans2016improved} & 0.96 ($\pm 0.07$) & - & - \\
{\it Conv-Ladder}~\cite{rasmus15} & \textbf{0.89} ($\pm 0.50$)& -  & -\\
\hline
{\it Conv-MMCVA} & 1.24 ($\pm0.54$) & \textbf{0.54} ($\pm0.04$) & \textbf{0.31}\\
\hline
\label{ssl-mnist-table}
\end{tabular}
\end{table}

\begin{figure}[!t]
\centering
\includegraphics[width=.8\columnwidth]{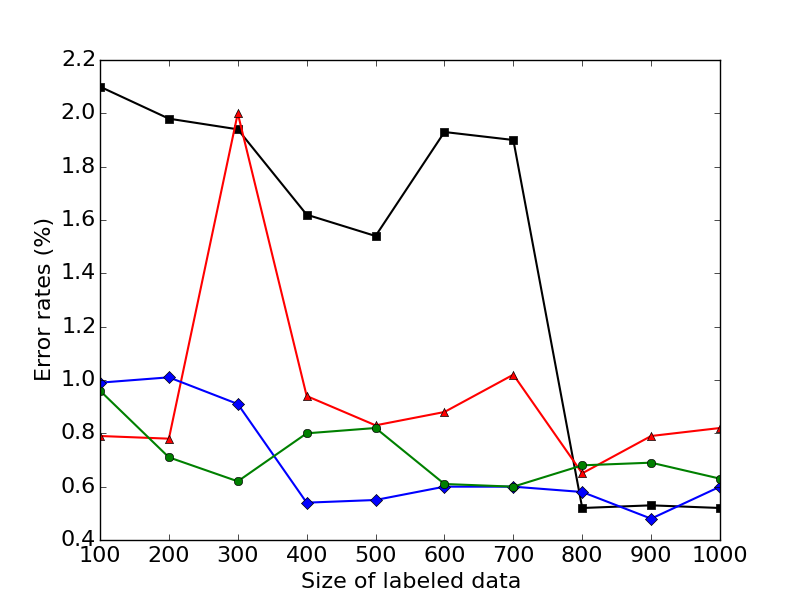}
\caption{Effect of size of labeled data on MNIST. The labeled data of smaller size is a subset of that of larger size in the same curve to reduce variance. Generally, the error rates decrease as the number of labels increase and the peaks may be caused by the poor quality of new added labeled data. Nevertheless, 800 labels are sufficient to achieve an error rate that is comparable to the supervised learning results of other DGMs.}
\label{fig:effect_labels}
\end{figure}

\begin{figure*}[!t]
\centering
\subfigure[MNIST data]{\includegraphics[width=0.49\columnwidth]{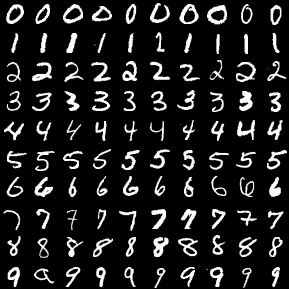}}
\subfigure[MNIST samples]{\includegraphics[width=0.49\columnwidth]{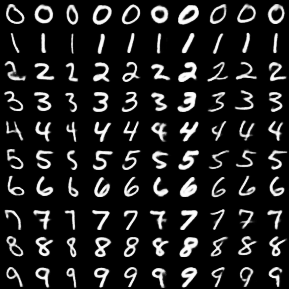}}
\subfigure[SVHN data]{\includegraphics[width=0.49\columnwidth]{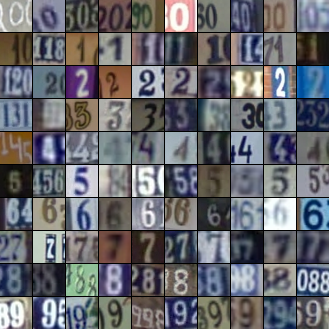}}
\subfigure[SVHN samples]{\includegraphics[width=0.49\columnwidth]{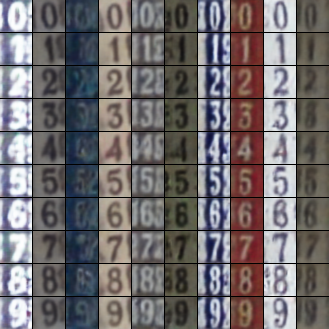}}
\caption{Class-conditional generation on MNIST (100 labels) and SVHN (1000 labels) datasets. (a) and (c) present 100 labeled training data sorted by class on MNIST and SVHN datasets respectively. (b) and (d) show samples on corresponding datasets where each row shares same class $y$ and each column shares same latent variables $\bbz$.}
\label{fig:disentangle}
\end{figure*}

We compare our Conv-MMCVA with a large body of previous methods on the MNIST dataset under different settings in Table~\ref{ssl-mnist-table}. Our method is competitive to the state-of-the-art results given 100 labels. As the number of labels increases, the max-margin principle significantly boosts the performance of Conv-MMCVA relative to the other models, including the Ladder Network~\cite{rasmus15}. Indeed, given 1,000 labels, Conv-MMCVA not only beats existing methods in the same setting, but also is comparable to the best supervised results of DGMs. The supervised learning results of Conv-MMCVA again confirm that by leveraging max-margin principle DGMs can achieve the same discriminative ability as the state-of-the-art CNNs with comparable architectures. We analyze the effect of the number of labels in Fig.~\ref{fig:effect_labels} for Conv-MMCVA, where the four curves share the same settings but use different random seeds to split data and initialize the networks.

\begin{table}[!t]
\renewcommand{\arraystretch}{1.2}
\caption{Error rates (\%) on SVHN and NORB datasets given 1000 labels.}
\centering
\begin{tabular}{lcc}
\hline
Algorithm & SVHN $n=1000$ & NORB $n=1000$\\
\hline
{\it M1+M2}~\cite{kingma14nips}  & 36.02 ($\pm0.10$) & 18.79 ($\pm0.05$)\\
{\it VAT}~\cite{miyato2015distributional}  & 24.63 & \textbf{9.88} \\
{\it ADGM}~\cite{maaloe16}  & 22.86 & 10.06($\pm0.05$)  \\
{\it SDGM}~\cite{maaloe16} &  16.61($\pm0.24$) & \textbf{9.40}($\pm0.04$) \\
\hline
{\it Improved-GAN}~\cite{salimans2016improved}  &\textbf{8.11} ($\pm1.3$)  & - \\
{\it Ensemble-10-GANs}~\cite{salimans2016improved}  &\textbf{5.88} ($\pm1.0$)  & - \\
\hline
{\it Conv-MMCVA} & \textbf{4.95} ($\pm0.18$) & \textbf{6.11} ($\pm0.58$) \\
\hline
\label{ssl-real-table}
\end{tabular}
\end{table}

Table~\ref{ssl-real-table} shows the classification results on the more challenging SVHN and NORB datasets. Following previous methods~\cite{kingma14nips,miyato2015distributional,maaloe16}, we use 1,000 labels on both datasets. We can see that our methods outperform the previous state-of-the-art substantially. {\it Ensemble-10-GANs} refers to an ensemble of 10 Improved-GANs~\cite{salimans2016improved} with 9-layer classifiers while we employ a single model with a shallower 6-layer classifier. Note that it is easy to further improve our model by using more advanced networks, e.g. ResNet~\cite{he2015deep}, without competition due to the separated architectures. In this paper, we focus on comparable architectures for fairness.

We further analyze the effect of the regularization terms to investigate
the possible reasons for the outstanding performance. If we omit the hat loss regularization, the Conv-MMCVA suffers from overfitting and only achieves 6.4\% error rates on the MNIST dataset given 100 labels. The underlying reason is that we approximate the full posterior inference by a greedy point estimation. If the prediction of the classifier is wrong, the generative model tends to interpret the unlabeled data with the incorrect label instead of enforcing the classifier to find the true label as in previous conditional DGM~\cite{kingma14nips}. However, the hat loss provides an effective way for the classifier to achieve a sufficiently good classification result, which can be fine-tuned according to the generative loss. In fact, trained to optimize the max-margin losses for both the labeled and unlabeled data, the classifier itself without the DGM can get 2.1\% error rates on MNIST given 100 labels. These results demonstrate the effectiveness of our proposed max-margin loss for the unlabeled data. We also reduce 0.2\% error rate in this setting by using the label-balance regularization. Besides the excellent performance, our Conv-MMCVA provides a potential way to apply class-conditional DGMs on large scale datasets with many more categories due to the efficient inference.

\subsubsection{Generative Performance}

\begin{figure}[!t]
\centering
\subfigure[NORB data]{\includegraphics[width=0.49\columnwidth]{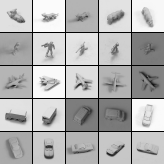}}
\subfigure[NORB samples]{\includegraphics[width=0.49\columnwidth]{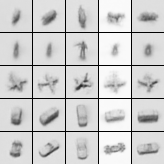}}
\caption{Class-conditional generation on NORB dataset (1000 labels). (a) and (b) are labeled training data and generated samples respectively.}
\label{fig:ssl-norb-sample}
\end{figure}

We demonstrate that our Conv-MMCVA has the ability to disentangle classes and styles given a small amount of labels on the MNIST, SVHN and NORB datasets, as shown in Fig.~\ref{fig:disentangle} and Fig.~\ref{fig:ssl-norb-sample}.
The images are generated by conditioning on a label $y$ and a style vector $\bbz$. On the MNIST and SVHN datasets, Conv-MMCVA are able to generate high-quality images and $\bbz$ can capture the intensities, scales and colors of the images.
Note that previous generation on SVHN in the semi-supervised learning setting is either unconditioned~\cite{salimans2016improved} or based on some preprocessed data~\cite{kingma14nips}.
Our samples are a little blurry on the NORB dataset, which contains elaborate images of 3D toys with different lighting conditions and points of view. Nevertheless, Conv-MMCVA can still separate these physical semantics from the general categories beyond digits. To the best of our knowledge, there is no competitive generative models to generate NORB data class-conditionally given the partially labeled data.

\section{Conclusions}

In this paper, we propose max-margin deep generative models (mmDGMs) and the class-conditional variants (mmDCGMs), which conjoin the predictive power
of max-margin principle and the generative ability of deep generative models. We develop a doubly
stochastic subgradient algorithm to learn all parameters jointly and consider two types of recognition
models with MLPs and CNNs respectively. We evaluate our mmDGMs and MMDCGMs in supervised learning and semi-supervised learning settings respectively. Given partially labeled data, we approximate the full posterior of the labels by a delta distribution for efficiency and propose additional max-margin and label balance losses for unlabeled data for effectiveness.

We present extensive results to demonstrate
that our methods can significantly improve the prediction performance of deep generative models,
while retaining the strong generative ability on generating input samples as well as completing
missing values. In fact, by employing CNNs in our mmDGMs and mmDCGMs, we achieve
low error rates on several datasets including MNIST, SVHN and NORB, which are competitive to the best fully
discriminative networks in supervised learning and improve the previous state-of-the-art semi-supervised results significantly.


%

\ifCLASSOPTIONcompsoc
  \section*{Acknowledgments}
\else
  \section*{Acknowledgment}
\fi

The work was supported by the National Basic Research
Program (973 Program) of China (No. 2013CB329403), National NSF of China (Nos. 61620106010, 61322308, 61332007), the Youth Top-notch Talent Support Program,
and Tsinghua TNList Lab Big Data Initiative.

\ifCLASSOPTIONcaptionsoff
  \newpage
\fi



%

\bibliography{bformat}
\bibliographystyle{plain}

%
\begin{IEEEbiography}[{\includegraphics[width=1in,height=1.2in,clip,keepaspectratio]{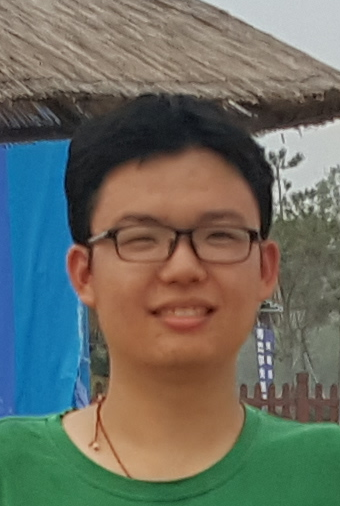}}]{Chongxuan Li}
received his BS from Institute for Interdiscriplinary Information Sciences, Tsinghua University, China. He is currently
working toward his PhD degree in
the Department of Computer Science and
Technology at Tsinghua University.
His research interests are primarily on statistical machine
learning, especially deep generative models for various learning tasks including unsupervised, (semi-)supervised and reinforcement learning.
\end{IEEEbiography}

\begin{IEEEbiography}[{\includegraphics[width=1in,height=1.2in,clip,keepaspectratio]{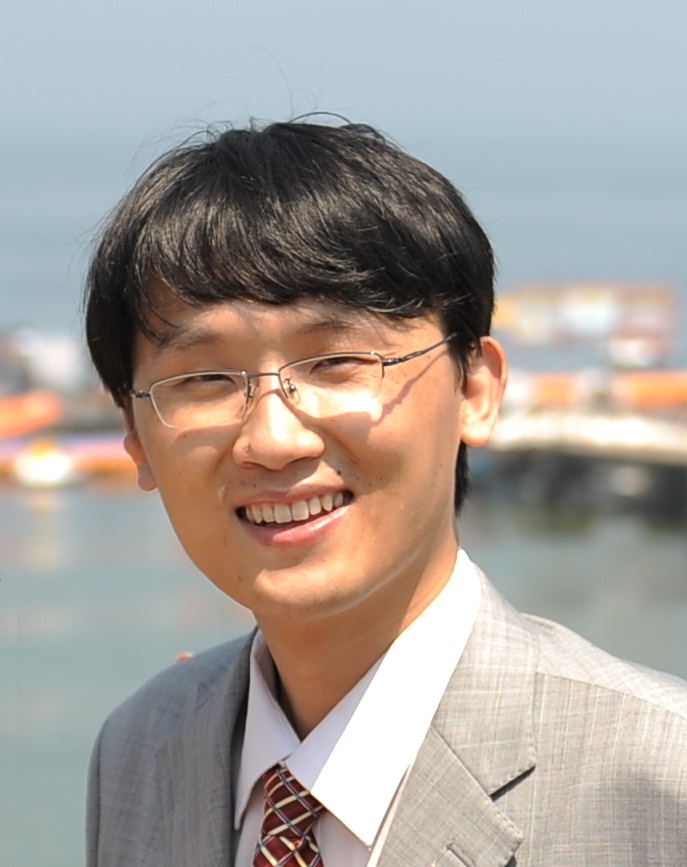}}]{Jun Zhu}
received his BS, MS and PhD
degrees all from the Department of Computer
Science and Technology in Tsinghua
University, China, where he is currently an
associate professor. He was a project scientist
and postdoctoral fellow in the Machine
Learning Department, Carnegie Mellon University.
His research interests are primarily
on developing statistical machine learning
methods to understand scientific and engineering
data arising from various fields. He
is a member of the IEEE.
\end{IEEEbiography}

\begin{IEEEbiography}[{\includegraphics[width=1in,height=1.2in,clip,keepaspectratio]{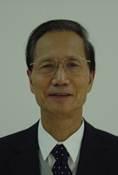}}]{Bo Zhang}
graduated from the Department
of Automatic Control, Tsinghua University,
Beijing, China, in 1958. Currently, he is a
Professor in the Department of Computer
Science and Technology, Tsinghua University
and a Fellow of Chinese Academy of
Sciences, Beijing, China. His main interests
are artificial intelligence, pattern recognition,
neural networks, and intelligent control. He
has published over 150 papers and four
monographs in these fields.
\end{IEEEbiography}

\newpage

\appendices

\section{Network Architecture of MMVA}
\label{app:architecture}

\begin{figure}[h]
\centering
\includegraphics[width=\columnwidth]{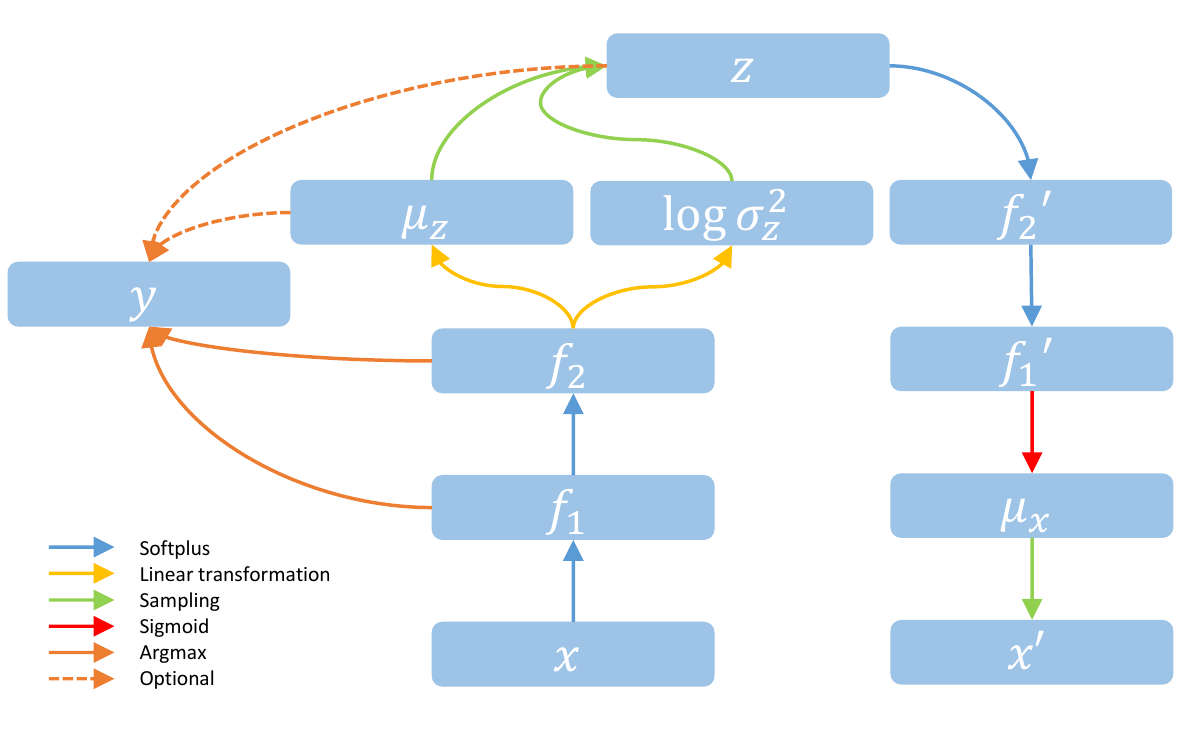}
\caption{Network architecture of MMVA with Gaussian hidden variables and Bernoulli visible variables.}
\label{architecture}
\end{figure}

We illustrate the network structure of MMVA with Gaussian hidden variables and Bernoulli visible variables in Fig.~\ref{architecture}.

\section{Manifold Visualization}
\label{app:manifold}

\begin{figure*}
\centering
\subfigure[VA]{\includegraphics[width=.8\columnwidth]{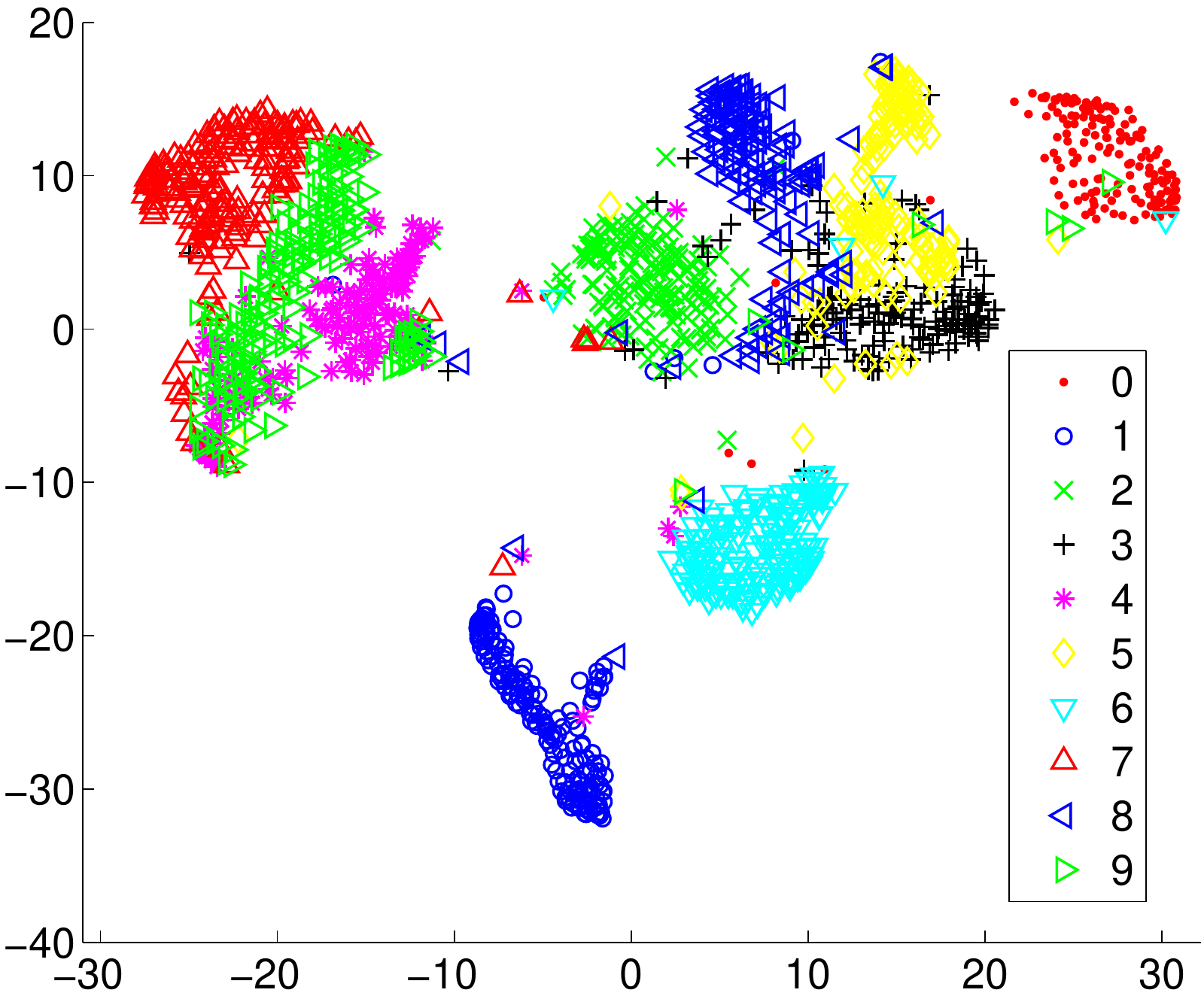}}
\subfigure[MMVA]{\includegraphics[width=.8\columnwidth]{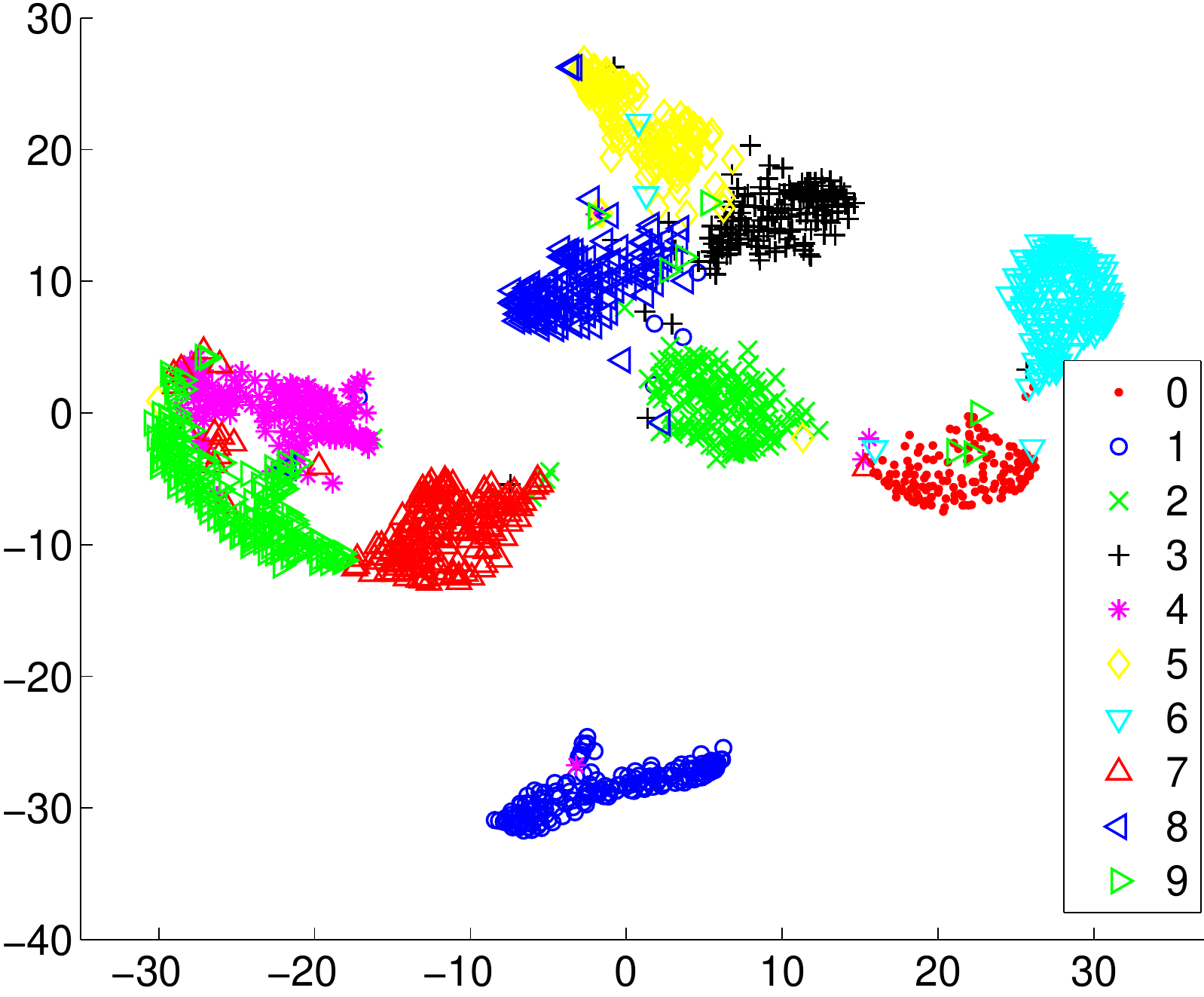}}
\caption{ t-SNE embedding results for both (a) VA and (b) MMVA.}
\label{embedding}
\end{figure*}

T-SNE embedding results of the features learned by VA and MMVA on 2D plane are shown in Fig.~\ref{embedding} (a) and Fig.~\ref{embedding} (b) respectively,
using the same data points randomly sampled from the MNIST dataset.
Compared to the VA's embedding,
MMVA separates the images from different categories better,
especially for the confusable digits such as digit ``4'' and ``9''. 
These results show that MMVA,
which benefits from the max-margin principle,
learns more discriminative representations of digits than VA.

\begin{figure*}[!t]
\centering
\subfigure[Original data ]{\includegraphics[width=0.49\columnwidth]{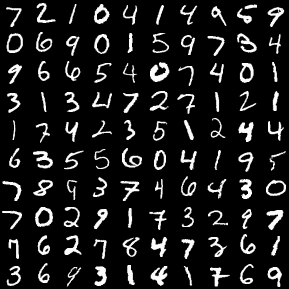}}
\subfigure[Noisy data]{\includegraphics[width=0.49\columnwidth]{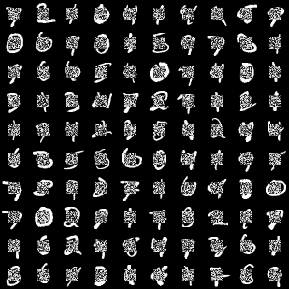}}
\subfigure[Results of Conv-VA]{\includegraphics[width=0.49\columnwidth]{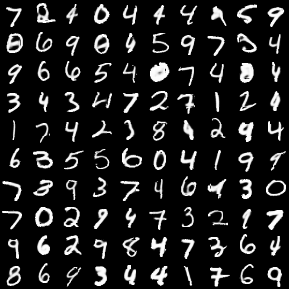}}
\subfigure[Results of Conv-MMVA]{\includegraphics[width=0.49\columnwidth]{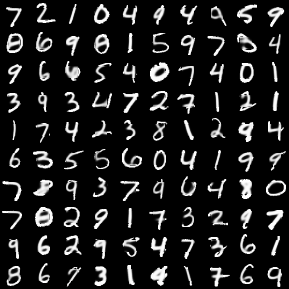}}
\caption{(a): original test data; (b) test data with missing value;
(c-d): results inferred by Conv-VA and Conv-MMVA respectively for 100 iterations.}
\label{cmmva-imputation}
\end{figure*}

\begin{figure*}
\centering
\subfigure[Original data ]{\includegraphics[width=0.49\columnwidth]{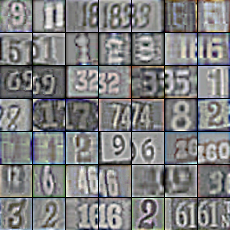}}
\subfigure[Noisy data]{\includegraphics[width=0.49\columnwidth]{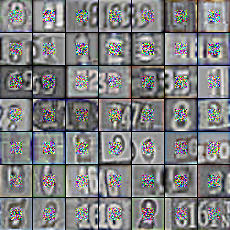}}
\subfigure[Results of Conv-VA]{\includegraphics[width=0.49\columnwidth]{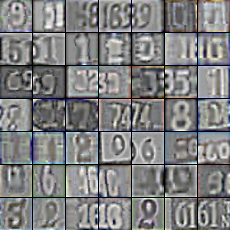}}
\subfigure[Results of Conv-MMVA]{\includegraphics[width=0.49\columnwidth]{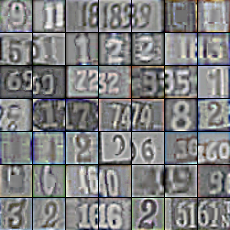}}
\caption{(a): original test data; (b) test data with missing value;
(c-d): results inferred by Conv-VA and Conv-MMVA respectively for 100 epochs.}
\label{svhn-imputation}
\end{figure*}

\section{Visualization of Imputation Results}
\label{app:vis_imputation}

The imputation results of Conv-VA and Conv-MMVA on MNIST and SVHN are shown in Fig.~\ref{cmmva-imputation} and Fig.~\ref{svhn-imputation} respectively.
On MNIST, Conv-MMVA makes fewer mistakes and refines the images better,
which accords with the MSE results as reported in the main text.
On SVHN, in most cases, Conv-MMVA could complete the images with missing values on this much harder dataset.
In the remaining cases, Conv-MMVA fails potentially due to the changeful digit patterns and less color contrast compared with hand-writing digits dataset.
Nevertheless, Conv-MMVA achieves comparable results with Conv-VA on inferring missing data.




\end{document}